\documentclass[preprint,12pt]{elsarticle}

\usepackage{amssymb}

\usepackage{caption}
\usepackage{tabularray}
\usepackage{url}
\usepackage{multirow}
\usepackage{longtable}
\usepackage{booktabs,multicol,multirow,tabularx,array} 
\usepackage{makecell}
\usepackage{float}
\usepackage{comment}
\usepackage{amsmath,amsfonts}
\usepackage{afterpage}
\usepackage[T1]{fontenc}
\usepackage{hyperref}
\usepackage{natbib}
\usepackage{graphicx}
\usepackage{color, soul}
\usepackage[subrefformat=parens,labelformat=parens]{subfig}
\usepackage{cleveref}
\usepackage{siunitx}
\usepackage{algorithm}
\usepackage[noend]{algpseudocode}
\makeatletter
\def\BState{\State\hskip-\ALG@thistlm}
\makeatother

\makeatletter 
\def\ps@pprintTitle{%
  \let\@oddhead\@empty
  \let\@evenhead\@empty
  \let\@oddfoot\@empty
  \let\@evenfoot\@oddfoot
}
\makeatother

\begin{document}

\begin{frontmatter}



\title{Step-by-Step Guidance to Differential Anemia Diagnosis with Real-World Data and Deep Reinforcement Learning}

%

\author[inst1,inst2]{Lillian Muyama\corref{cor1}}

\affiliation[inst1]{organization={Inria Paris},
            city={Paris},
            postcode={75013}, 
            country={France}}

\author[inst4]{Estelle Lu}
\author[inst3]{Geoffrey Cheminet}
\author[inst4]{Jacques Pouchot}
\author[inst4]{Bastien Rance}
\author[inst4]{Anne-Isablle Tropeano}
\author[inst1,inst2,inst3]{Antoine Neuraz}
\author[inst1,inst2]{Adrien Coulet}
\cortext[cor1]{Corresponding author: \url{lillian.muyama@inria.fr}}

\affiliation[inst2]{organization={Centre de Recherche des Cordeliers, Inserm, Université Paris Cité, Sorbonne Université},
            city={Paris},
            postcode={75006}, 
            country={France}}

\affiliation[inst4]{organization={Hôpital Européen Georges Pompidou, Assistance Publique - Hôpitaux de Paris},
            city={Paris},
            postcode={75015}, 
            country={France}}
            
\affiliation[inst3]{organization={Hôpital Necker, Assistance Publique - Hôpitaux de Paris},
            city={Paris},
            postcode={75015}, 
            country={France}}

\begin{abstract}
Clinical diagnostic guidelines outline the key 
questions to answer to reach 
a diagnosis. 
Inspired by guidelines, we aim to develop a model that learns 
from electronic health records to determine
the optimal sequence of actions for accurate diagnosis. 
Focusing on anemia and its sub-types, we employ deep reinforcement learning (DRL) algorithms and evaluate their performance on both a synthetic dataset, which is based on expert-defined diagnostic pathways, and a real-world dataset. 
We investigate the performance of these algorithms across various scenarios.
Our experimental results demonstrate that DRL algorithms perform competitively with state-of-the-art methods while offering the significant advantage of progressively generating pathways to the suggested diagnosis, providing a transparent decision-making process that can guide and explain diagnostic reasoning.

\end{abstract}




\end{frontmatter}



\section{Introduction}\label{sec:intro}
Clinicians often rely on Clinical Practice Guidelines (CPGs) to make informed diagnostic decisions. CPGs are systematically developed recommendations to assist healthcare providers in choosing suitable healthcare strategies tailored to specific clinical scenarios \cite{field1990clinical}.  The primary purpose of these guidelines is to standardize and rationalize clinical decisions. For diagnostic purposes, the recommended actions may include physical examinations of a patient, laboratory tests, imaging studies, and other relevant assessments.

However, despite their undeniable role in clinical decision-making, CPGs have several limitations. First, their development requires input from a diverse panel of experts, leading to a 
time-consuming and costly process. This implies that updates to CPG take several years, which makes them unsuitable for rapidly evolving medical practices, such as the introduction of new diagnostic tests or the emergence of a new disease. Second, due to the substantial resources needed to develop these guidelines, it is not practical to create CPGs for every known medical condition. Third, guidelines are generally designed to address the needs of the majority of the population, potentially overlooking rare conditions or uncommon
populations. Additionally, the adoption of CPGs in real-world clinical settings is often complicated by various factors, such as outdated recommendations, clinicians' reliance on personal knowledge and experience over guidelines, and resource constraints. In response to these challenges, several studies have explored innovative approaches for learning clinical pathways from data. These methods have the potential to complement guidelines by providing insights where guidelines are absent, incomplete, or unsuitable.

The vast amount of data collected during care, 
particularly through Electronic Health Records (EHRs), presents an 
opportunity to enhance clinical decision-making. EHRs include diverse patient information, including laboratory test results, medications, examinations, and diagnoses, reflecting clinical practice. Following directions proposed by \cite{adler2021next}, our study
develops a data-driven methodology that supports the clinical decision-making process in a step-by-step manner.
We believe this approach could reduce unnecessary tests, optimize costs, and offer 
personalized and accurate diagnoses, particularly for uncommon 
and overlooked
patients.

In this work, we build on a previous study \cite{muyama2024deep}, where authors used a Deep Reinforcement Learning (DRL) approach to learn diagnostic decision pathways for anemia and systemic lupus erythematosus.  
The authors created synthetic datasets, which were labeled based on existing CPGs, and manually added noise and missing data to test the robustness of their approach. 
In the present study, we  
go further, but focus only on anemia,
for three main reasons: its diagnosis is primarily based on a series of laboratory tests available in most EHRs; it is a common diagnosis, implying that the associated amount of data is sufficient to train ML models; and the differential diagnosis of anemia is often complex, making its guidance particularly useful.

Our approach proceeds as follows: First, we collaborated with a 
clinician specialized in internal medicine to develop a diagnostic decision tree (DT) for anemia, which we name the ``expert-defined'' DT. This tree served as the basis for generating a synthetic dataset, on which we applied our method to generate baseline diagnostic pathways.
Second, we 
tested our approach on a real-world dataset, 
under three distinct scenarios: 
(\textit{i}) applying the model trained solely on synthetic data to the real-world data;
(\textit{ii}) fine-tuning the synthetic data-trained model with a subset of the real-world data; 
and (\textit{iii}) training a new model from scratch using a subset of the real-world dataset. 

Our main contributions are: 
1) We demonstrate that deep reinforcement learning methods have the potential to provide step-by-step guidance for clinical diagnosis in a real-world setting.
2) We show that in cases of insufficient data, the use of a synthetic dataset may improve model performance.

\section{Related Work}\label{sec:related_work}
The discovery of clinical pathways from patient data has been widely explored using mostly unsupervised learning techniques such as process mining and topic modeling. For instance, \cite{bakhshi2023optimizing} employed process mining to discover clinical pathways for sepsis, 
\cite{vogt2018applyig} applied K-medoids and logistic regression to identify pathways to reduce hospitalizations, and \cite{xu2016summarizing} combined process mining with topic modeling to generate clinical pathways for intracranial hemorrhage. Other frequently employed unsupervised learning methods include hidden Markov models \cite{huang2018probabilistic, najjar2018two} and sequence mining \cite{hur2020facilitating, dagliati2018care}. 
However, these approaches often require that the trajectory (sequence of clinical events) of each patient is known, which can be problematic in real-world settings where event timings may overlap, be delayed, or recorded unreliably.

Additionally, several studies have leveraged machine learning methods to aid clinical decision-making by predicting patient outcomes \cite{koshimizu2020prediction, miotto2016}, diagnoses \cite{lipton2015learning, choi2016doctor}, among other tasks.
However, unlike these works, our study aims not only to provide diagnostic predictions but also to detail the steps taken to reach that diagnosis, thereby mimicking the decision-making process of a clinician.

Previous studies have applied reinforcement learning (RL) methods for costly feature acquisition in classification tasks \cite{li2021active, janisch2019classification}. However, these works did not specifically focus on clinical diagnosis.
RL Methods have also been utilized to develop optimal dosing strategies in various domains, including chemotherapy \cite{huo2022multi, maier2021reinforcement}, radiotherapy \cite{gallagher2023learning, moreau2021reinforcement}, and glucose control \cite{zhu2020basal, liu2020deep}. These applications can be viewed as developing \emph{treatment} pathways for patients.

Several works \cite{tang2016inquire, wei2018task, kao2018context} built dialogue systems, where patients were queried about their symptoms, eventually leading to a diagnosis.
While our approach similarly formulates the diagnosis process as a sequential decision-making problem using DRL, these studies rely on self-reported symptoms. 
In contrast, we aim to use EHRs, which we believe offer a robust alternative as they encompass data routinely collected in clinical practice, including objective and standardized measurements such as laboratory results. 

In another related work \cite{yu2023deep}, the authors sought to optimize the financial cost of several clinical processes, including the prediction of Acute Kidney Injury using DRL. While they were successful in reducing laboratory test costs compared to other models, the sequences of actions leading to these reductions and subsequent diagnoses were not discussed or shown. This omission neglects the relevance and explainable dimension of pathways.



\section{Methods}\label{sec:methods}
\subsection{Decision Problem}\label{sec:decision_problem}
We consider the clinical diagnosis process as a sequential decision-making problem and formulate it as a Markov Decision Process (MDP) \cite{Littman2001markov}, following the RL \cite{sutton2018reinforcement} paradigm. 
Accordingly, we define an \emph{agent} that interacts with an \emph{environment} to maximize a cumulative reward signal. 
At each time step $t$, the agent receives an observation $o$ of the \emph{state} of the environment, takes an \emph{action}, and obtains a \emph{reward}.  
This sequence of interactions between the agent and the environment, from the initial state to the terminal state, is called \emph{an episode}.
The goal is to learn a policy (\textit{i.e.}, a function that maps states to actions) that maximizes the reward signal. 

Let $\mathcal{D}$ denote a dataset of size $n \times (m+1)$ composed of $n$ patients, $m$ features, and one diagnosis. $F$ is the set of the $m$ feature names; and $C$ the set of possible diagnosis values. An instance $D^i$ in $\mathcal{D}$ is a pair $(X^i, Y^i)$ with $X^i = \{x^i_1,\dots, x^i_j, \dots , x^i_m\}$ where $x^i_j$ is the value of the feature $j$ for the patient $i$, $m$ is the total number of features and $Y^i \in C$ is the anemia diagnosis of patient $i$.
Accordingly, our MDP is defined by the quadruple  $(S, A, T, R)$  as follows:
\begin{itemize}
    \item $S$ is the set of states.
    At each time step, the agent receives an observation $o$ of the state $s_t$, which is a vector of fixed size $m$ comprising the values of the features that the agent has already queried at time $t$.
   Features that have not been queried yet are associated with the value -1. Equation \ref{eqn:state_eqn} 
     defines the $j^{th}$ element of $o$, where $F'$ denotes the set of features that the agent has already queried. A simulation of a step in the anemia environment is presented in Algorithm \mbox{\ref{env_step}}, and an illustration of how the state is updated is shown in Table \mbox{\ref{tab:env_state}}, both in \mbox{\ref{apd:env}}.
   \begin{equation}
    \label{eqn:state_eqn}
      o_j = \begin{cases}
      \;x_j, & \text{if $f_j \in F' $}\\
      -1, & \text{otherwise.}
    \end{cases}
    \end{equation}
    
    \item $A$ is the set of possible actions, which is the union of the set of \emph{feature value acquisition actions}, $A_f$ (or \emph{feature actions} for short), and the set of 
    \emph{diagnostic actions}, $A_d$. 
     At each time step, the agent takes an action $a_t \in A$. 
     Actions from $A_f$ involve selecting values from the set of features $F$. Accordingly, a specific value $f_j \in F$ will trigger the action of querying the Clinical Data Warehouse (CDW) for the value of this feature.
     Actions from $A_d$ involve selecting values from the set of possible diagnoses $C$. 
     At any time step $t$, if $a_t \in A_d$, the episode is terminated. Also, for \textit{anemia}, if an episode reaches the specified maximum number of steps without reaching a diagnosis, it is terminated.
    
    \item $T$ denotes the transition function that specifies the probability of transitioning from a state $s_t$ to a state $s_{t+1}$ given an agent action $a_t \in A$.

    \item $R$ is the reward function. $r_{t+1}$, which can be written as $R(s_t, a_t)$, is the immediate reward when an agent in state $s_t$ takes action $a_t$. For a diagnostic action $a_t \in A_d$, if the diagnosis is correct, the reward is +1, 
    otherwise, the reward is -1. 
    \begin{equation}
    \label{eqn:diag_actions}
       \text{if}\ a_t \in A_d, R(s_t, a_t) = \begin{cases}
      \;\;\,1, & \text{if $a_t = Y^i $}\\
      -1, & \text{otherwise}
    \end{cases}
    \end{equation}

For a feature action $a_t \in A_f$, if the feature has already been queried, the agent incurs a penalty of -1, and the episode is terminated. 
Otherwise, the agent receives a penalty of $-\frac{1}{2n}$ where $n$ is the number of features in the dataset. Thus, the reward function for the feature actions is formalized as follows: 
\begin{equation}
     \label{eqn:anem_feat_actions}
      \text{if}\ a_t \in A_f, R(s_t, a_t) = \begin{cases}
      -1, & \text{if $a_t \in F' $}\\
      -\frac{1}{2n}, & \text{otherwise}.
    \end{cases}
    \end{equation}
\end{itemize}

\subsection{Deep Q-Network}
\textbf{Q-learning} \cite{watkins1992q} is an RL algorithm that determines the best action to take in a given state based on the expected future reward of taking that action in that particular state. This expected future reward is referred to as the Q-value of that state-action pair,
denoted as $Q(s,a)$. 
During model training, the Q-values are updated using the Bellman Equation as described in equation \ref{eqn:q_learning}.

In our study, given the large and continuous state space, we propose using a \textbf{Deep Q-Network (DQN)} \cite{mnih2015human}, which employs a neural network to approximate the Q-value function. Based on a previous study \cite{muyama2024deep}, we used \textbf{Dueling DQN} and \textbf{Dueling Double DQN (DDQN)} paired with \textbf{Prioritized Experience Replay (PER)} for our experiments. A more detailed description of these methods is presented in \ref{apd:dqn}.

\subsection{Datasets}\label{sec:data}
\subsubsection{Synthetic Dataset}
\begin{figure}
    \centering
    \includegraphics[scale=0.2]{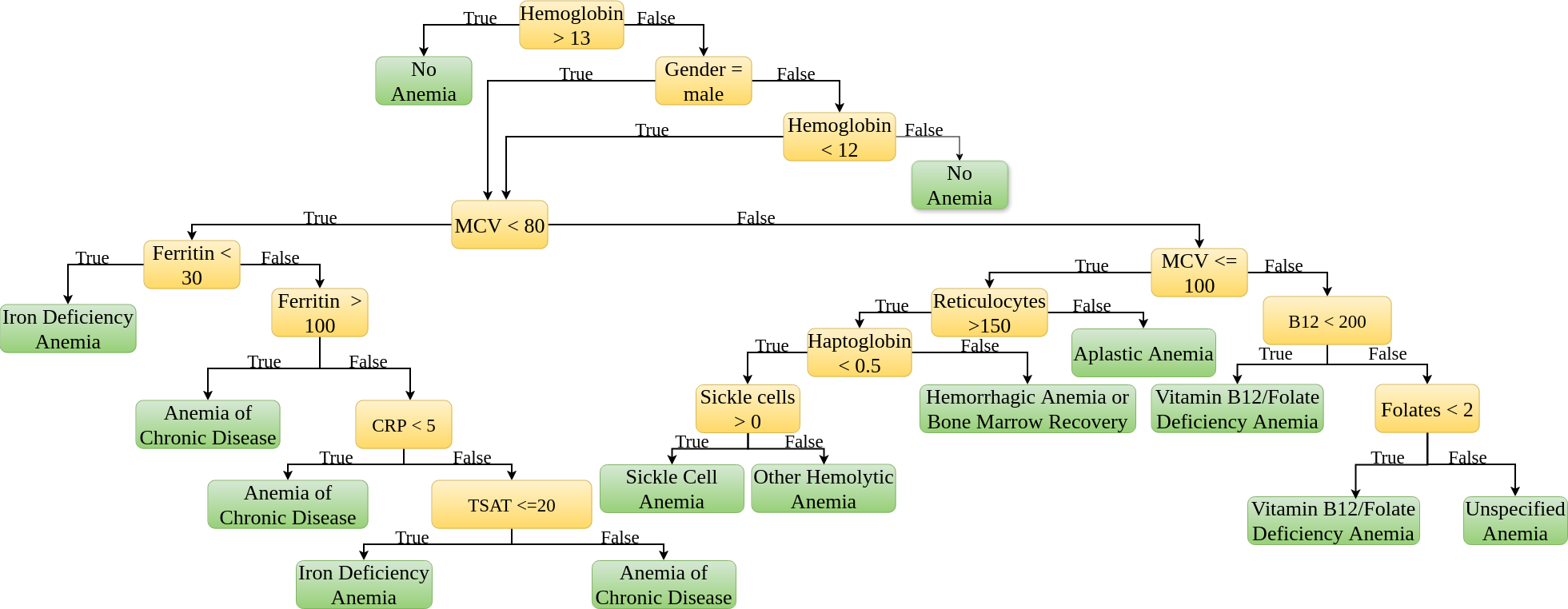}
    \caption{The expert-defined decision tree used to label the synthetic dataset.}
    \label{fig:tree}
\end{figure}

We 
created a synthetic dataset following the 
methodology described in \cite{muyama2024deep}. However, we refined the decision tree they proposed 
based on expert input from a qualified internist ensuring that the diagnostic logic closely reflects both the local clinical practice 
and the data encoding of the local hospital. 
The resulting 
decision tree, shown in Figure~\ref{fig:tree}, was used to label the synthetic dataset.

It consists of 69,879 instances, each associated with a single label (anemia diagnosis); and 11 features pertinent to anemia diagnosis: \emph{hemoglobin, gender, Mean Corpuscular Volume (MCV), ferritin, C-Reactive Protein (CRP), Transferrin Saturation (TSAT), reticulocytes, haptoglobin, Vitamin B12, folate, and sickle cells.}
Details about this dataset are shared in \ref{apd:syn_dataset}.

\subsubsection{Real-World Dataset (RWD)}

The RWD for this study was sourced from the CDW of Georges Pompidou European hospital in Paris, France. It includes patients diagnosed with anemia between January 2018 and December 2023, who were admitted to the hospital for the first time, either directly or indirectly through the emergency department, to the Internal Medicine or Geriatrics departments. These departments were chosen due to their significant volume of anemia cases.
The patient inclusion was carried out in two phases. 

In the first phase, we included patients with an ICD-10 code for anemia recorded up to 10 days after admission, along with their lab results from the first 15 days of admission, and their demographics. If multiple lab results existed in this timeframe, we selected the earliest one.

In the second phase, we identified additional patients whose clinical notes included an anemia diagnosis term.
Using Medkit \cite{neuraz2024medkit}, a Natural Language Processing Python library, we processed hospitalization and consultation notes, focusing on the ``reason for hospitalization'' and ``conclusion'' sections, to detect specific anemia diagnosis-related terms, as provided by a clinical expert. 
Once these notes were identified, we retrieved the corresponding lab results recorded up to 15 days after admission. Lab results were only included if their timestamps were either before or on the same date as the clinical note containing the anemia diagnosis. If multiple lab results were available in the timeframe, we only kept the closest to the timestamp of the relevant clinical note.

For the \emph{No anemia} class, we randomly chose a quarter of the patients from the two departments from the same time period who had no ICD-10 code for anemia within the first 10 days after admission, and no clinical notes mentioning anemia within 15 days after admission.

For the final dataset, we excluded patients without recorded hemoglobin values, 
and those with diagnoses not represented in the expert-defined DT (Figure \ref{fig:tree}). Due to insufficient data, the ``Hemorrhagic anemia or bone marrow recovery'' class was also excluded.
In total, the RWD consists of 1,127 patients with the same features as those of the synthetic dataset. More details about this dataset are found in \ref{apd:rwd_dataset}. A list of the ICD-10 codes used for the data extraction process from the CDW, as well as the anemia terms used for Medkit can be found at \url{https://github.com/lilly-muyama/RWD-Anemia-Diagnostic-Pathways}.



\subsection{Experimental design}
We propose four experiments for this study, which are described in this section.

In the first experiment, we allocate 80\% of the synthetic dataset for training and 20\% for testing. Additionally, 10\% of the training dataset (\textit{i.e.} 8\% of the total dataset) is used for validation. We use the Dueling DQN-PER and Dueling DDQN-PER models, trained on this dataset, to generate diagnostic pathways.

For the remaining experiments, we split the RWD into 80\% for training, 10\% for validation, and 10\% for testing. This split is chosen due to the smaller size of the hospital dataset, and the presence of classes with very few samples ($n <$ 20). By ensuring that the validation set is of equal size and a similar class distribution to the test set, we aim to achieve a well-balanced and representative evaluation of model performance.

In the second experiment, we train the same two DRL models used in the first experiment, but exclusively on the RWD.
In the third experiment, we apply the models trained on the synthetic dataset from the first experiment directly to the RWD test set, without further training or modification, to generate diagnostic pathways.
In the final experiment, we employ transfer learning by fine-tuning the models previously trained on the synthetic dataset using the RWD training set. These fine-tuned models are then applied to generate diagnosis pathways for the RWD test set.

Additionally, for comparison purposes, we included a random agent that makes decisions randomly, and a tree-based agent, which acts according to the expert-defined tree shown in Figure~\ref{fig:tree}.

\subsection{Evaluation Approach and Implementation}
We report \emph{F1} and \emph{ROC-AUC} scores using a one-vs-rest approach and macro-averaging. We further computed the \emph{accuracy} as the ratio of episodes that terminated with a correct diagnosis, and the \emph{mean episode length (MEL)} as the average number of actions performed for each episode. 
Ten runs were conducted for each experiment, with each run using a different seed, which influences the initial weights of the models.

All models were trained for a fixed number of timesteps, with checkpoints saved at regular intervals. The checkpoint with the best performance, determined by a specified metric (accuracy/F1/ROC-AUC) evaluated on the validation data, was selected.

We implemented our environment using the OpenAI Gym Python library \cite{brockman2016openai}. 
To build our agents, we used the stable-baselines package \cite{stable-baselines}. The hyperparameter values, listed in Table \ref{tab:hyperparams} in \ref{apd:dqn}, were selected based on prior knowledge, existing literature, and experimentation. 
We used the default values specified in the stable-baselines package documentation for the unlisted remaining hyperparameters. The source code of these experiments is available at \url{https://github.com/lilly-muyama/RWD-Anemia-Diagnostic-Pathways}.

To evaluate the identification of anemia terms in the clinical notes with our matching algorithm, we constructed a small manually annotated set of clinical notes ($n$ = 100) on which we tested our
approach.

\section{Results}\label{sec:results}


\subsection{Results using the Synthetic Dataset}
In our initial round of experiments, we employed Dueling DQN-PER and Dueling Double DQN-PER on the synthetic dataset.
Each experiment was conducted ten times using different seeds, with the results presented in Table \ref{tab:synthetic_results}.
Here, the tree-based agent
achieved a perfect score since it follows the expert-described DT used to label the dataset.
The results reported are for the DQN models selected based on their F1 score.

\begin{table}[!htbp]
\centering
\resizebox{\textwidth}{!}{ 
    \begin{tabular}{|l|l|l|l|l|}
    \hline
    \bfseries Model & \bfseries Accuracy & \bfseries F1 & \bfseries ROC-AUC & \bfseries MEL\\\hline
        Random agent & 10.08 ± 0.32 & 11.14 ± 0.38 & 49.98 ± 0.21 & 2.02 ± 0.01 \\ \hline
        Tree-based agent & \textbf{100.00 ± 0.00} & \textbf{100.00 ± 0.00 } & \textbf{100.00 ± 0.00}  & 4.79 ± 0.00 \\ \hline
         Learned decision tree & 99.97 ± 0.01 & 99.97 ± 0.01 & 99.98 ± 0.00 & 7.85 ± 0.00 \\ \hline
        Dueling DQN-PER & 98.98 ± 0.19 & 99.23 ± 0.15 & 99.46 ± 0.10 & 4.47 ± 0.07 \\ \hline
        Dueling DDQN-PER & 99.02 ± 0.21 & 99.25 ± 0.15 & 99.47 ± 0.11 & 4.41 ± 0.04 \\\hline
    \end{tabular}
    }
    \caption{Performance of models on the synthetic dataset.}
     \label{tab:synthetic_results}%
\end{table}

\subsection{Results using the Real-World Dataset}
Next, we conducted experiments using the RWD. 
The results of the second experiment, where models were trained exclusively on the RWD, are presented in Table \ref{tab:from_scratch}. 
While the accuracy score was consistent across all models, the F1 scores demonstrated high variability, as indicated by their standard deviations. Among the models, Dueling DQN-PER based on the F1 score performed the best in this set of experiments.

Table \ref{tab:as_is} displays the results of the experiments where models trained on the synthetic dataset were applied directly to the RWD test set, without further training or modification
The models were selected based on their F1 scores, specifically those closest to the mean F1 score reported in Table \ref{tab:synthetic_results}.
For comparison, we also included a decision tree trained on the synthetic dataset, which was also used to generate pathways for the RWD test set.
The results reveal
that while the decision tree had the best accuracy, the Dueling DDQN-PER model had the best F1 and ROC-AUC scores. The decision tree also had on average, pathways twice as long as the DQN models.

Lastly, Table \ref{tab:finetuned} shows the results of the experiments with the fine-tuned synthetic dataset-trained models (used in Table \ref{tab:as_is}).
The results
show an improvement in performance, especially for the dueling DQN-PER model. 

In Table \ref{tab:sota_results}, we present a summary of the results and compare the best-performing DQN agent in terms of F1 score (the fine-tuned Dueling DQN-PER) against other agents.
To assess the quality of diagnoses made, we also compared the models' performance to Random Forest (RF), Extreme Gradient Boosting (XGBoost), and a Feed-Forward Neural Network (FFNN) even though these algorithms do not generate pathways.
The hyperparameter values for these models were determined using the validation set.
RF achieved the best accuracy, while the fine-tuned dueling DQN-PER Model had the best F1 and XGBoost had the best ROC-AUC score.

\begin{table}[!h]
  \centering
  \resizebox{\textwidth}{!}{ 
    \begin{tabular}{|l|l|l|l|l|}
    \hline
    \bfseries Model & \bfseries Accuracy & \bfseries F1 & \bfseries ROC-AUC & \bfseries MEL\\\hline
       Random agent & 8.60 ± 4.03 & 5.91 ± 3.18 & 46.73 ± 6.40 & 2.02 ± 0.12 \\ \hline
        Tree-based agent & 57.02 ± 0.00 & 27.9 ± 0.00 & 69.75 ± 0.00 & 3.18 ± 0.00 \\ \hline
        RWD-trained DT & 81.93 ± 0.45 & 55.82 ± 0.94 & 75.47 ± 0.50 & 10.72 ± 0.00 \\ \hline
        Random Forest & \textbf{86.84 ± 0.93} & 57.29 ± 1.36 & 75.69 ± 0.56 & N/A \\ \hline
        XGBoost & 84.21 ± 0.00 & 61.77 ± 0.00 & \textbf{78.61 ± 0.00} & N/A  \\ \hline
        FFNN & 75.44 ± 2.65 & 45.34 ± 5.92 & 70.99 ± 2.65 & N/A \\ \hline
        Best DQN (Fine-tuned Dueling DQN-PER) & 77.19 ± 0.00 & \textbf{63.48 ± 0.00}  & 77.67 ± 0.00  & 3.19 ± 0.00 \\ \hline
    \end{tabular}
    }
    \caption{Performance of models on the real-world dataset. Random Forest, XGBoost, and FFNN do not generate pathways and therefore do not have a mean episode length.}
    \label{tab:sota_results}%
\end{table}


\section{Discussion}\label{sec:discussion}
\subsection{Anemia Term Identification}
Our matching algorithm achieved a precision score of 86\% in identifying relevant anemia-related terms within the small, manually annotated clinical notes dataset. Upon analyzing the misdetections, the primary cause was uncertain diagnoses, accounting for 42.9\% of errors. In these cases, while the anemia term was correctly detected in the appropriate section of the clinical note, it did not represent a confirmed diagnosis.
Another 28.6\% of misdetections resulted from the anemia term being identified outside the two relevant sections, as the algorithm failed to recognize that it belonged to a different part of the note.
Additional reasons for misdetections included negation (21.4\%), and instances where the patient carried the gene but did not have the disease (7.1\%).
These misdetections may have influenced the overall results, and enhancing the precision of the matching algorithm could lead to improvements in our approach to diagnostic pathway generation. 


\subsection{DQN Performance with synthetic dataset}
As presented in Table \ref{tab:synthetic_results}, the DQN models exhibited slightly lower performance than the learned decision tree.
For instance, when examining the pathways and diagnoses generated by one of the Dueling DDQN-PER models, the class with the lowest precision score was Vitamin B12/Folate deficiency anemia, with a precision of 98\%. Further analysis revealed that all misdiagnosed cases were patients with \emph{Unspecified anemia}, incorrectly diagnosed as having \emph{Vitamin B12/Folate deficiency anemia}.
As shown in Figure~\ref{fig:tree}, these two classes occupy the same branch of the diagnosis tree.
Upon further examination of the most common pathway of these misdiagnosed episodes, which was \textit{``hemoglobin $\rightarrow$ MCV $\rightarrow$ Folate $\rightarrow$ Vitamin B12/Folate deficiency anemia''}, we found that the folate levels in these cases was near the threshold value in the tree. Specifically, their mean folate value was 2.00 $\mu$g/L (SD=0.01). 
This trend was consistent across other classes, where misdiagnosed cases often had values near diagnostic thresholds, resulting in either a false diagnosis or an \emph{Inconclusive diagnosis} in cases with missing values.
From a clinical perspective, this is acceptable as lab results near thresholds may not lead to a definitive diagnosis.

\subsection{DQN Performance with real-world dataset}
Overall, fine-tuning the Dueling DQN-PER model resulted in a noticeable improvement. When analyzing the diagnoses made by the best-performing model (based on F1 score), two classes in particular, \textit{i.e.} \emph{Sickle cell anemia} and \emph{Other hemolytic anemia} showed a remarkable improvement.
Specifically, for \emph{Sickle cell anemia}, the recall increased from 9\% with the synthetic dataset-trained DQN Model to 65\% with the RWD-trained model. After fine-tuning the synthetic dataset-trained model, recall improved further to 83\%. 
In clinical settings where patient diagnosis is especially critical, achieving a high recall score is essential. 
Furthermore, for the \emph{Other hemolytic anemia} class, while both the synthetic dataset-trained and RWD-trained DQN models had a precision and recall of 0\%, the fine-tuned model showed improved performance, with precision and recall scores of 67\% and 29\%, respectively. Other classes also exhibited improved performance, but the limited data prevented us from drawing definitive conclusions.

\subsection{DQN \textit{vs.} expert-defined DT}
Notably, the expert-defined DT achieved a precision and recall score of 0 for the \emph{sickle cell anemia} class. In contrast, the DQN achieved a precision of 79\% and a recall of 83\% for this class. 
Further analysis revealed that most \emph{sickle cell anemia} cases were assigned an \emph{inconclusive diagnosis} by the expert-defined tree agent. In examining the most common pathway for these cases, which was ``hemoglobin $\rightarrow$ MCV $\rightarrow$ Ferritin $\rightarrow$ Inconclusive diagnosis'', we found that the cases following this pathway had a mean MCV level of 71.86 fL (SD=4.49), and therefore according to the tree (see Figure~\ref{fig:tree}), the next step would be to request for the ferritin level. However, since the ferritin values were absent in these cases, an \emph{inconclusive diagnosis} was given.
On the other hand, the DQN correctly diagnosed 6 of these 7 cases, likely due to its exposure to similar cases during training.

\subsection{Generated Pathways}
\begin{figure}[htbp]
    \centering
    \includegraphics[width=\linewidth]{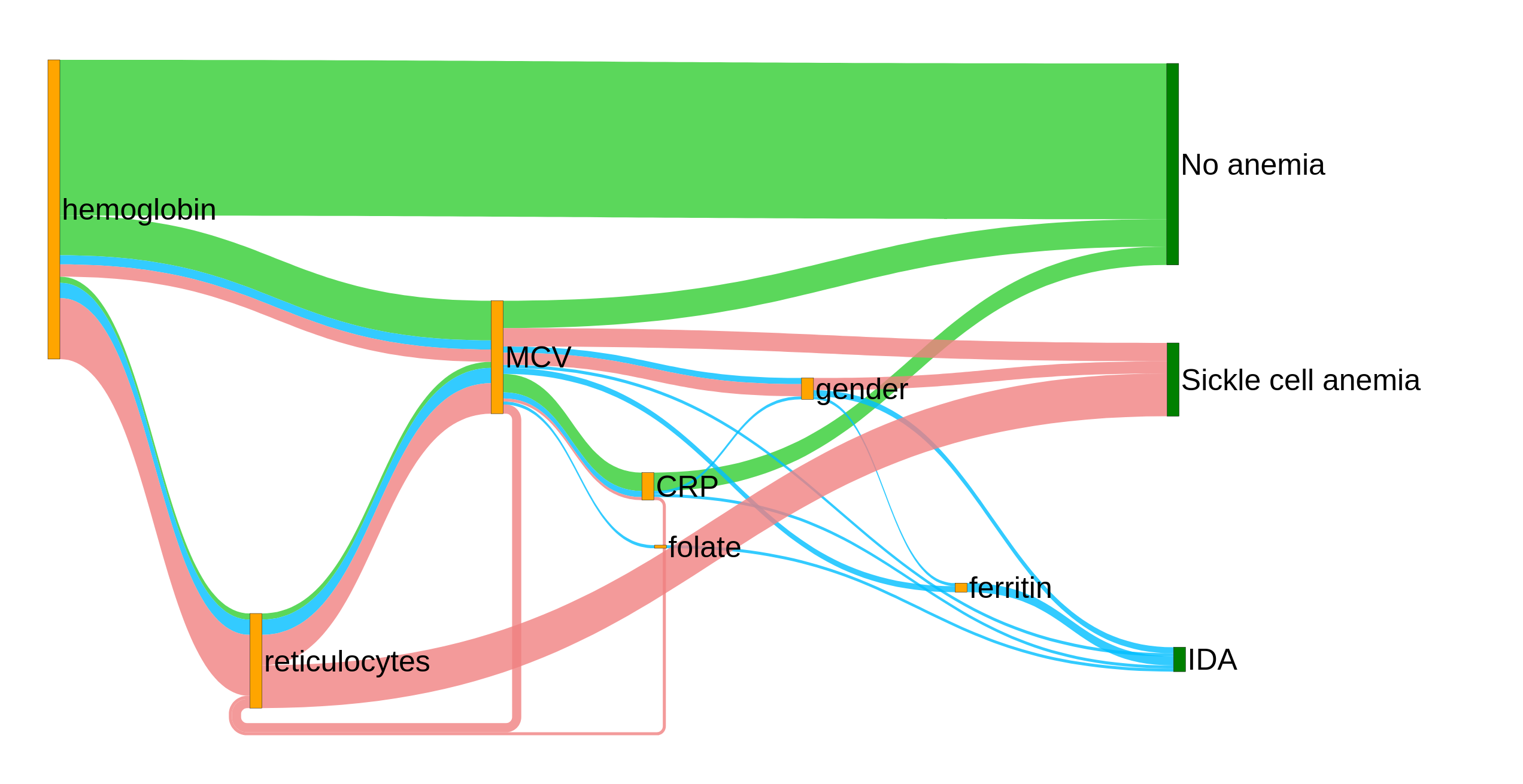}
    \caption{Diagnostic pathways generated by the fine-tuned dueling DQN-PER model.}
    \label{fig:finetuned_pathways}
\end{figure}

Figure~\ref{fig:finetuned_pathways} illustrates the pathways generated for the \emph{No anemia} and \emph{Sickle cell anemia} classes by the DQN model that achieved the highest F1 score, whose performance metrics are presented in Table \ref{tab:sota_results}.
Additional pathways generated by models in the different scenarios can be found in Figures \ref{fig:modified_dt_pathway} to \ref{fig:finetuned_pathway} in \ref{apd:pathways}.
Algorithm \ref{testing} in \ref{apd:env} describes the process of generating
pathways using the test dataset.

Diagnostic pathways were not generated for certain classes by some models. For example, \emph{Unspecified anemia} had no pathways from the expert-defined DT or the synthetic dataset-trained model, and \emph{Anemia of chronic disease (ACD)} had none from the RWD-trained and fine-tuned DQN models. However, these two classes had only a single instance in the RWD test dataset making it difficult to draw any definitive conclusions. Moreover, expert opinion suggests that these two classes are traditionally challenging to diagnose, as their diagnosis may be more complex than indicated by the expert-defined DT.

An intriguing finding is that the expert-defined DT generated no pathways for \emph{Sickle cell anemia}. Additionally, the synthetic-trained DQN model diagnosed only two cases despite \emph{Sickle cell anemia} being the second most common class in the dataset. This indicates that the expert-defined DT may be inadequate for diagnosing this condition and needs revision to better reflect real-world clinical practice. In contrast, DQN models trained or fine-tuned on the RWD showed a significant improvement in diagnosing \emph{Sickle cell anemia}.

To assess pathway similarity across models, we calculated the average Levenshtein distance between patient pathways generated by each model. 
Pathways were encoded as strings, with each feature in the pathway represented by a unique character. We then computed the average Levenshtein distance for pathways across model pairs. 
The closest similarity was between the expert-defined DT and the synthetic dataset-trained model, with an average Levenshtein distance of 0.29 (SD=0.73). 
The second most similar pathways were between the RWD-trained DQN model and the fine-tuned DQN model, with a mean Levenshtein distance of 0.92 (SD=1.23), 
while the greatest dissimilarity was between the synthetic dataset-trained and fine-tuned models, with a mean distance of 1.48 (SD=1.35).

Finally, we examined a patient case with the most dissimilar pathways (mean Levenshtein distance of 3.33). The actual diagnosis for this patient, \emph{Sickle cell anemia}, was identified only by the RWD-trained DQN model, while other models' pathways led to an \emph{Inconclusive diagnosis}. Upon further analysis, it was noted that this patient had an MCV value of 59 fL, which falls significantly below the threshold for diagnosing \emph{Sickle cell anemia} according to the expert-defined DT in Figure \ref{fig:tree}. Consequently, both the expert-defined DT and the synthetic dataset-trained model proceeded to test for \emph{ferritin} as the next feature in the pathway and thus failed to reach the correct diagnosis. 
On the other hand, the RWD-trained model likely encountered similar cases during training, which enabled it to generate a pathway leading to a correct diagnosis. However, the fine-tuned model resulted in an \emph{Inconclusive diagnosis} due to missing haptoglobin data.

\subsection{Limitations}
One limitation of our study is the size of the real-world dataset and the class imbalance. While some classes were well-represented with hundreds of samples, others had far fewer samples as illustrated in Figure~\ref{fig:rwd_class_distribution}. Consequently, we were unable to draw definitive conclusions on the performance of the models and the generated pathways for these particular classes, and this may also have led to the significant variability in the performance metrics, especially the F1 score, in the experiments using only the real-world dataset.
Additionally, the dataset was sourced from a hospital that is a reference center for Sickle cell disease, which may account for the relatively high number of \emph{Sickle cell anemia} cases in the dataset, making it the second-most represented class after \emph{No anemia}.

Additionally, training the DQN Models required considerably more time and computational resources compared to other agents, as shown in Table \ref{tab:computing_time}. However, once the model is trained, generating diagnostic pathways for test instances is straightforward, since the policy has already been learned by the model.

\section{Conclusion}\label{sec:conclusion}

In this work, we applied DRL methods to generate clinical diagnostic pathways for patients using EHR data. These methods were applied to two datasets: a synthetic dataset generated from an expert-defined decision tree, and a real-world dataset sourced from a local hospital.
We explored different scenarios and demonstrated that DRL algorithms have the potential to guide, step-by-step, the clinical diagnostic process by generating diagnostic pathways that can provide insight into decisions. These pathways have the important advantage of guiding the diagnosis of individual patients, and also for summarizing alternate diagnostic routes for a specific patient population.

In future work, we plan to extend our approach to a broader range of medical conditions, especially those where diagnosis is not based on a decision tree. Additionally, since laboratory tests are usually ordered as panels, this can be incorporated as well. Furthermore, it will be interesting to apply our approach to diagnoses that rely on multimodal data, potentially acquired over extended time periods, rather than solely on laboratory test results.

\section*{Acknowledgements}\label{sec:acknowledgement}
This work is supported by the Inria CORDI-S Ph.D. program, and benefited from a government grant managed by the Agence Nationale de la Recherche under the France 2030 program, reference ANR-22-PESN-0007 ShareFAIR, and ANR-22-PESN-0008 NEUROVASC.


\newpage
\bibliographystyle{elsarticle-num} 
\bibliography{references}

\begin{thebibliography}{10}
\expandafter\ifx\csname url\endcsname\relax
  \def\url#1{\texttt{#1}}\fi
\expandafter\ifx\csname urlprefix\endcsname\relax\def\urlprefix{URL }\fi
\expandafter\ifx\csname href\endcsname\relax
  \def\href#1#2{#2} \def\path#1{#1}\fi

\bibitem{field1990clinical}
M.~J. Field, K.~N. Lohr, et~al., Clinical practice guidelines, Directions for a new program (1990) 1990.

\bibitem{adler2021next}
J.~Adler-Milstein, J.~H. Chen, G.~Dhaliwal, \href{https://doi.org/10.1001/jama.2021.22396}{{Next-Generation Artificial Intelligence for Diagnosis: From Predicting Diagnostic Labels to “Wayfinding”}}, JAMA 326~(24) (2021) 2467--2468.
\newblock \href {https://doi.org/10.1001/jama.2021.22396} {\path{doi:10.1001/jama.2021.22396}}.
\newline\urlprefix\url{https://doi.org/10.1001/jama.2021.22396}

\bibitem{muyama2024deep}
L.~Muyama, A.~Neuraz, A.~Coulet, Deep reinforcement learning for personalized diagnostic decision pathways using electronic health records: A comparative study on anemia and systemic lupus erythematosus, arXiv preprint arXiv:2404.05913 (2024).

\bibitem{bakhshi2023optimizing}
A.~Bakhshi, E.~Hassannayebi, A.~H. Sadeghi, Optimizing sepsis care through heuristics methods in process mining: A trajectory analysis, Healthc. Anal. 3 (2023) 100187.

\bibitem{vogt2018applyig}
V.~Vogt, S.~M. Scholz, L.~Sundmacher, Applying sequence clustering techniques to explore practice-based ambulatory care pathways in insurance claims data, Eur. J. Public Health 28~(2) (2018) 214--219.

\bibitem{xu2016summarizing}
X.~Xu, T.~Jin, J.~Wang, Summarizing patient daily activities for clinical pathway mining, in: 2016 IEEE 18th international conference on e-health networking, applications and services (Healthcom), IEEE, 2016, pp. 1--6.

\bibitem{huang2018probabilistic}
Z.~Huang, Z.~Ge, W.~Dong, K.~He, H.~Duan, Probabilistic modeling personalized treatment pathways using electronic health records, J. Biomed. Inform. 86 (2018) 33--48.

\bibitem{najjar2018two}
A.~Najjar, D.~Reinharz, C.~Girouard, C.~Gagn{\'e}, A two-step approach for mining patient treatment pathways in administrative healthcare databases, Artif. Intell. Med. 87 (2018) 34--48.

\bibitem{hur2020facilitating}
C.~Hur, J.~Wi, Y.~Kim, Facilitating the development of deep learning models with visual analytics for electronic health records, Int. J. Environ. Res. Public Health 17~(22) (2020) 8303.

\bibitem{dagliati2018care}
A.~Dagliati, V.~Tibollo, G.~Cogni, L.~Chiovato, R.~Bellazzi, L.~Sacchi, Careflow mining techniques to explore type 2 diabetes evolution, J. Diabetes Sci. Technol. 12~(2) (2018) 251--259.

\bibitem{koshimizu2020prediction}
H.~Koshimizu, R.~Kojima, K.~Kario, Y.~Okuno, Prediction of blood pressure variability using deep neural networks, International Journal of Medical Informatics 136 (2020) 104067.

\bibitem{miotto2016}
R.~Miotto, L.~Li, B.~A. Kidd, J.~T. Dudley, Deep patient: an unsupervised representation to predict the future of patients from the electronic health records, Scientific Reports 6~(1) (2016) 1--10.

\bibitem{lipton2015learning}
Z.~C. Lipton, D.~C. Kale, C.~Elkan, R.~Wetzel, Learning to diagnose with {LSTM} recurrent neural networks, arXiv preprint arXiv:1511.03677 (2015).

\bibitem{choi2016doctor}
E.~Choi, M.~T. Bahadori, A.~Schuetz, W.~F. Stewart, J.~Sun, {Doctor AI}: Predicting clinical events via recurrent neural networks, in: Machine learning for healthcare conference, PMLR, 2016, pp. 301--318.

\bibitem{li2021active}
Y.~Li, J.~Oliva, Active feature acquisition with generative surrogate models, in: International Conference on Machine Learning, PMLR, 2021, pp. 6450--6459.

\bibitem{janisch2019classification}
J.~Janisch, T.~Pevn{\`y}, V.~Lis{\`y}, Classification with costly features using deep reinforcement learning, in: Proceedings of the AAAI Conference on Artificial Intelligence, Vol.~33, 2019, pp. 3959--3966.

\bibitem{huo2022multi}
L.~Huo, Y.~Tang, Multi-objective deep reinforcement learning for personalized dose optimization based on multi-indicator experience replay, Appl. Sci. 13~(1) (2022) 325.

\bibitem{maier2021reinforcement}
C.~Maier, N.~Hartung, C.~Kloft, W.~Huisinga, J.~de~Wiljes, Reinforcement learning and {B}ayesian data assimilation for model-informed precision dosing in oncology, CPT Pharmacometrics Syst. Pharmacol. 10~(3) (2021) 241--254.

\bibitem{gallagher2023learning}
K.~Gallagher, M.~Strobl, R.~Gatenby, P.~Maini, A.~Anderson, Learning to adapt-deep reinforcement learning in treatment-resistant prostate cancer, bioRxiv (2023) 2023--04.

\bibitem{moreau2021reinforcement}
G.~Moreau, V.~Fran{\c{c}}ois-Lavet, P.~Desbordes, B.~Macq, Reinforcement learning for radiotherapy dose fractioning automation, Biomedicines 9~(2) (2021) 214.

\bibitem{zhu2020basal}
T.~Zhu, K.~Li, P.~Herrero, P.~Georgiou, Basal glucose control in type 1 diabetes using deep reinforcement learning: An in silico validation, IEEE J. Biomed. Health Inform. 25~(4) (2020) 1223--1232.

\bibitem{liu2020deep}
Z.~Liu, L.~Ji, X.~Jiang, W.~Zhao, X.~Liao, T.~Zhao, S.~Liu, X.~Sun, G.~Hu, M.~Feng, et~al., A deep reinforcement learning approach for type 2 diabetes mellitus treatment, in: 2020 IEEE International Conference on Healthcare Informatics (ICHI), IEEE, 2020, pp. 1--9.

\bibitem{tang2016inquire}
K.-F. Tang, H.-C. Kao, C.-N. Chou, E.~Y. Chang, Inquire and diagnose: Neural symptom checking ensemble using deep reinforcement learning, in: NIPS Workshop on Deep Reinforcement Learning, 2016.

\bibitem{wei2018task}
Z.~Wei, Q.~Liu, B.~Peng, H.~Tou, T.~Chen, X.-J. Huang, K.-F. Wong, X.~Dai, Task-oriented dialogue system for automatic diagnosis, in: Proceedings of the 56th Annual Meeting of the Association for Computational Linguistics (Volume 2: Short Papers), 2018, pp. 201--207.

\bibitem{kao2018context}
H.-C. Kao, K.-F. Tang, E.~Chang, Context-aware symptom checking for disease diagnosis using hierarchical reinforcement learning, in: Proceedings of the AAAI Conference on Artificial Intelligence, Vol.~32, 2018.

\bibitem{yu2023deep}
Z.~Yu, Y.~Li, J.~Kim, K.~Huang, Y.~Luo, M.~Wang, Deep reinforcement learning for cost-effective medical diagnosis, arXiv preprint arXiv:2302.10261 (2023).

\bibitem{Littman2001markov}
M.~Littman, Markov decision processes, in: N.~J. Smelser, P.~B. Baltes (Eds.), International Encyclopedia of the Social \& Behavioral Sciences, Pergamon, Oxford, 2001, pp. 9240--9242.

\bibitem{sutton2018reinforcement}
R.~S. Sutton, A.~G. Barto, Reinforcement learning: An introduction, MIT press, 2018.

\bibitem{watkins1992q}
C.~J. Watkins, P.~Dayan, Q-learning, Machine learning 8~(3) (1992) 279--292.

\bibitem{mnih2015human}
V.~Mnih, K.~Kavukcuoglu, D.~Silver, A.~A. Rusu, J.~Veness, M.~G. Bellemare, A.~Graves, M.~Riedmiller, A.~K. Fidjeland, G.~Ostrovski, et~al., Human-level control through deep reinforcement learning, Nature 518~(7540) (2015) 529--533.

\bibitem{neuraz2024medkit}
A.~Neuraz, G.~Vaillant, C.~Arias, O.~Birot, K.-T. Huynh, T.~Fabacher, A.~Rogier, N.~Garcelon, I.~Lerner, B.~Rance, A.~Coulet, \href{https://arxiv.org/abs/2409.00164}{Facilitating phenotyping from clinical texts: the medkit library} (2024).
\newblock \href {http://arxiv.org/abs/2409.00164} {\path{arXiv:2409.00164}}.
\newline\urlprefix\url{https://arxiv.org/abs/2409.00164}

\bibitem{brockman2016openai}
G.~Brockman, V.~Cheung, L.~Pettersson, J.~Schneider, J.~Schulman, J.~Tang, W.~Zaremba, Open{AI} {G}ym, arXiv preprint arXiv:1606.01540 (2016).

\bibitem{stable-baselines}
A.~Hill, A.~Raffin, M.~Ernestus, A.~Gleave, A.~Kanervisto, R.~Traore, P.~Dhariwal, C.~Hesse, O.~Klimov, A.~Nichol, M.~Plappert, A.~Radford, J.~Schulman, S.~Sidor, Y.~Wu, Stable {B}aselines, \url{https://github.com/hill-a/stable-baselines} (2018).

\bibitem{van2016deep}
H.~Van~Hasselt, A.~Guez, D.~Silver, Deep reinforcement learning with double {Q}-learning, in: Proceedings of the AAAI conference on artificial intelligence, Vol.~30, 2016.

\bibitem{wang2016dueling}
Z.~Wang, T.~Schaul, M.~Hessel, H.~Hasselt, M.~Lanctot, N.~Freitas, Dueling network architectures for deep reinforcement learning, in: International conference on machine learning, PMLR, 2016, pp. 1995--2003.

\bibitem{schaul2015prioritized}
T.~Schaul, J.~Quan, I.~Antonoglou, D.~Silver, Prioritized experience replay, arXiv preprint arXiv:1511.05952 (2015).

\end{thebibliography}

\appendix
\counterwithin{figure}{section}
\counterwithin{table}{section}

\renewcommand{\thetable}{\Alph{section}.\arabic{table}}
\renewcommand{\thefigure}{\Alph{section}.\arabic{figure}} 

\newpage
\section{The environment}\label{apd:env}

We implemented our environment using the OpenAI Gym Python library. During training, the environment is reset using a random instance from the dataset at the beginning of each episode. The state of the environment is initialized as a vector of size 17, filled with -1s, corresponding to the set of feature actions, $A_f$ described in Section \mbox{\ref{sec:decision_problem}}.

Algorithm \mbox{\ref{env_step}} provides a simulation of a step by an agent in the environment for the \emph{anemia} use case. In this algorithm, $i$ represents the training instance used to reset the environment, $s_t$ is the state of the environment, $Y^i$ is the diagnosis of $i$, $action$ is the action selected by the agent, $r_t$ is the reward, and \emph{done} indicates whether an episode is completed.
Table \mbox{\ref{tab:env_state}} illustrates how the state of the environment is updated when a \emph{feature action} is selected by the agent.

\begin{algorithm}
    \caption{Environment step for episode $i$ in state $s_t$}\label{env_step}
    \begin{algorithmic}[0]
    \Function{Step}{$action$}
    \State $epLen = epLen+1$ \Comment{Increase episode length}
    \If {$action \in A_d$}   
        \If{$action$ == $Y_i$}
            \State $r_t = 1$   \Comment{Reward if diagnosis is correct}
        \Else 
            \State $r_t = -1$   \Comment{Penalty if diagnosis is wrong}
        \EndIf
        \State done = True   \Comment{Terminate the episode}
    \ElsIf {$epLen == maxEpLen$}
        \State $r_t = -1$   \Comment{Penalty if max length is reached}
        \State done = True   \Comment{Terminate the episode}
    \ElsIf {$action \in F'$}
        \State $r_t = -1$   \Comment{Penalty if action has already been performed}
        \State done = True   \Comment{Terminate the episode}
    \Else 
        \State $r_t = 0$
        \State done = False
        \State $F'.add(action)$   \Comment{Add action to list of selected features}
        \State value = $fetchValue(action)$ \Comment{Fetch feature value from CDW}
        \State $s_t[action]$ = value \Comment{Update the state}
    \EndIf
    \State \textbf{return} $s_t$, $r_t$, done , $F'$
    \EndFunction
        
    \end{algorithmic}
\end{algorithm}

\begin{table}[]
    \centering
    \begin{tabular}{|c|c|c|c|c|}
    \hline
         \textbf{Step} & \textbf{State} & \textbf{Action} & \textbf{Value} & \textbf{Updated State}  \\
         \hline
         \rule{0pt}{3ex} 0 & [-1, -1, -1, ...., -1, -1] & - & - & [-1, -1, -1, ..., -1, -1] \\
         \hline
         \rule{0pt}{3ex} 1 & [-1, -1, -1, ...., -1, -1] & $a^0_f$ & 0.2 & [0.2, -1, -1, ..., -1, -1] \\
         \hline
         \rule{0pt}{3ex} 2 & [0.2, -1, -1, ...., -1, -1] & $a^{15}_f$ & 72 & [0.2, -1, -1, ..., 72, -1] \\
         \hline
    \end{tabular}
   
    \caption{Illustration of how the state is updated when a \emph{feature action} is selected. The state is represented as a vector of fixed size 17, and the possible feature actions range from $a_f^0$ to $a_f^{16}$.}
    \label{tab:env_state}
\end{table}

Algorithm \mbox{\ref{testing}} shows the process of generating pathways for patients in the test set using the trained model.

\begin{algorithm}
\caption{Generating pathways for test set data}\label{testing}
\begin{algorithmic}[0]
    \State \textbf{Input:} Test dataset $D_{test}$; Trained model $M$
    \State \textbf{Output:} List of Pathways $P$
    \State Instantiate empty pathway list: $P$ 
    \State Instantiate test environment: $env$ = Env$(D_{test})$ 
    \For{\textbf{all} $\textbf{$i$} \in D_{test}$}
        \State Instantiate empty pathway $p_i$
        \State Reset test environment using test instance: $obs$ = $env$.reset($i$)
        \State done = False
        \While{not done}
            \State $action$ = $M$.selectAction(obs)
            \State obs, rew, done = $env$.STEP($action$)
            \State $p_i$ = $p_i$.add($action$)
        \EndWhile
        \State $P$ = $P$.add($p_i$)
    \EndFor
\end{algorithmic}
\end{algorithm}

\section{Deep Q-Network}\label{apd:dqn}
\subsection{Model definitions}
\textbf{Q-learning} \cite{watkins1992q} is an RL algorithm that determines the best action to take in a given state based on the expected future reward for that action in that state. This expected future reward is known as the Q-value of that state-action pair, 
denoted as $Q(s,a)$. 
At each time step, the agent selects an action according to a policy $\pi$, with the goal of finding the optimal policy $\pi^*$ that maximizes the reward function. 
During model training, the Q-values are updated using the Bellman Equation as follows:

\begin{align}
\label{eqn:q_learning}
Q(s_t, a_t) \leftarrow & Q(s_t, a_t) + \alpha[r_{t+1} +\gamma\ \underset{a}{\max}\  Q(s_{t+1}, a) -\nonumber\\
  &\quad Q(s_t, a_t)]
\end{align}

where $\alpha$ is the learning rate, $r_{t+1}$  is the reward received after taking action $a_t$ in state $s_t$, and $\gamma$ is the discount factor that determines the importance of future reward relative to immediate reward.

In our use case, due to the large state space, we propose using a \textbf{Deep Q-Network} \cite{mnih2015human} which employs a neural network to approximate the Q-value function.
A DQN consists of two networks with similar architecture: the policy network, which interacts with the environment and learns the optimal policy, and the target network, which is used to define the target Q-value. At each time step, the policy network takes a state $s_{t}$ as its input and outputs the Q-values for taking the different actions in that state. 
The DQN algorithm uses an $\epsilon$-greedy strategy, where a random action is selected with probability $\epsilon$ (exploration), and the action with the highest estimated Q-value with probability 1-$\epsilon$ (exploitation).
The weights of the target network are frozen and updated at a specified interval by copying the weights of the policy network. The DQN algorithm learns by minimizing the loss function shown in Equation \ref{eqn:dqn_loss_fn}, where $\theta$ and $\theta^-$ represent the weights of the policy and target networks, respectively. Additionally, at each time step, a record of the model's interaction with the environment, known as an experience, is stored in a memory buffer in the form $(s_t, a_t, r_{t+1}, s_{t+1})$.

\begin{equation}
    \label{eqn:dqn_loss_fn}
    L(\theta) = \mathop{\mathbb{E}}[(r_{t+1} + \gamma\ \underset{a}{\max}\  Q(s_{t+1}, a, \theta^-) - Q(s_t, a_t, \theta))^2]
\end{equation}

To improve DQN stability and performance, several extensions of the DQN algorithm have been developed and are utilized in this paper. We briefly describe these extensions below:

\textbf{Double DQN} \cite{van2016deep}: 
In Double DQN, the action is selected using the policy network, while the target network estimates the value of that action. This is in contrast to the standard DQN where the same network is used for both tasks. The loss function is thus modified as follows:

\begin{align}
\label{eqn:ddqn_loss_fn}
 L(\theta) &= \mathop{\mathbb{E}}[(r_{t+1} + \gamma\ Q(s_{t+1}, \underset{a}{\arg\max}\ Q(s_{t+1}, a, \theta), \theta^-) \nonumber\\
  &\quad - Q(s_t, a_t, \theta))^2]
\end{align}

\textbf{Dueling DQN} \cite{wang2016dueling}: In Dueling DQN, the Q-value function is split into two parts: a value function $V(s)$ that provides the value for being in a given state, and an advantage function $A(s,a)$ that gives the advantage of taking action $a$ in state $s$, as compared to other actions. These two functions are then combined to compute the Q values as shown in Equation \ref{eqn:dueling_dqn}.

\begin{equation}
\label{eqn:dueling_dqn}
    Q(s,a) = V(s) + (A(s,a) - \frac{1}{|\mathcal{A}|} \underset{a}\sum A(s,a))
\end{equation}

where $|\mathcal{A}|$ denotes the total number of possible actions.

\textbf{Prioritized Experience Replay} \cite{schaul2015prioritized}: In PER, each experience stored in the replay buffer is assigned a priority based on its estimated importance. Experiences with higher priorities are sampled more frequently during training compared to experiences with lower priorities. This approach aims to prioritize and replay experiences that contribute more significantly to learning and improving the model's performance.

\subsection{Model hyperparameters}
Table \ref{tab:hyperparams} shows the hyperparameter values for the models used in this study.

\begin{table}[!ht]
\centering
    \begin{tabular}{|p{3.5cm}|p{2.35cm}|p{2.35cm}|p{2.35cm}|p{3.7cm}|}
    \hline
    \textbf{Hyperparameter} & \textbf{Value} \\ \hline
       Buffer size & 1000000 \\ 
        Learning rate & 0.0001 \\
        Target network update frequency & 10000 \\ 
        Learning starts & 50000 \\
        Final epsilon value & 0.05 \\
        Discount factor & 0.99 \\
        Train frequency & 4\\ \hline
    \end{tabular}
    \caption{DQN hyperparameter values.}
    \label{tab:hyperparams}
\end{table}

\section{The datasets}\label{apd:datasets}
\subsection{Synthetic dataset}\label{apd:syn_dataset}
\begin{figure}[htbp]
    \centering
    \includegraphics[width=\linewidth]{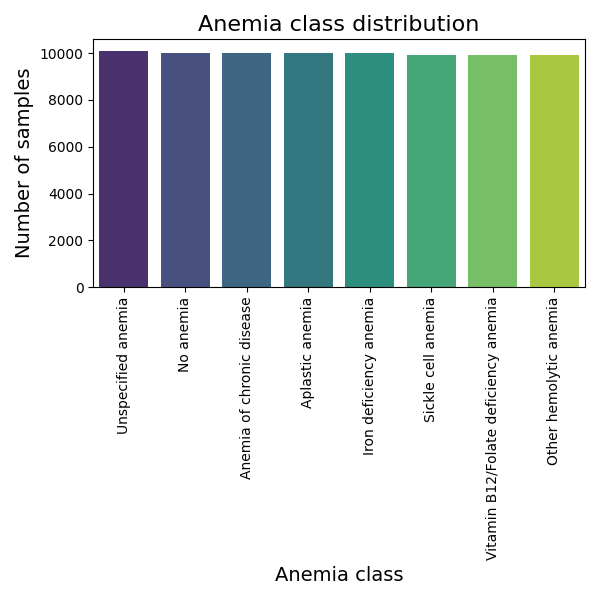}
    \caption{Class distribution in the synthetic dataset.}
    \label{fig:syn_class_distribution}
\end{figure}

\begin{figure}[htbp]
    \centering
    \includegraphics[width=\linewidth]{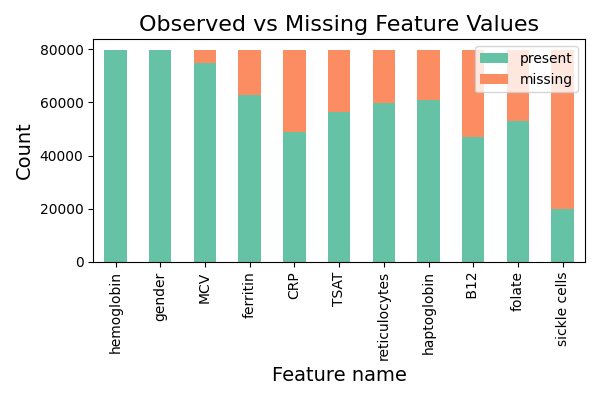}
    \caption{Observed \textit{vs.} missing values for each feature in the synthetic dataset.}
    \label{fig:syn_missing}
\end{figure}

Figure~\ref{fig:syn_class_distribution} shows the class distribution of the synthetic dataset, Figure~\ref{fig:syn_missing} depicts the ratio of observed versus missing values for each feature, Table \ref{tab:syn_instance} illustrates an example instance from the dataset, and Table \ref{tab:syn_summary_stats} presents the summary statistics of the dataset.

\begin{table}[htbp]
\centering
    \begin{tabular}[]{@{}|l|l|@{}}
    \hline
         \bfseries Feature & \bfseries Value \\\hline
         \textbf{Hemoglobin} & 10.62 g/100mL\\
         \textbf{Gender} & Male \\
         \textbf{MCV} & 82.50 fL\\
         \textbf{Ferritin} &  587.61 $\mu$g/L\\
         \textbf{CRP} & 276.05 mg/L\\
         \textbf{TSAT} & 43.58\% \\
         \textbf{Reticulocytes} & 22.34 G/L\\
         \textbf{Haptoglobin} & 2.38 g/L\\
         \textbf{Vitamin B12} & 1211.11 ng/L\\
         \textbf{Folate} & 18.29 $\mu$g/L\\
         \textbf{Sickle cells} & - \\         
         \textbf{Label} & Aplastic anemia\\\hline
    \end{tabular}
  \caption{An instance in the synthetic dataset. ``-'' represents a missing value.}
  \label{tab:syn_instance}
\end{table}

\begin{table}[htbp]
    \fontsize{7pt}{7pt}\selectfont
    \centering    
    \begin{tabular}[]{@{}|l|l|l|l|@{}}

  \hline
    \bfseries Feature & \bfseries All Classes & \bfseries No anemia & \bfseries \makecell{Vitamin B12/Folate \\ deficiency anemia} \\ \hline
    \textbf{Hemoglobin} & 8.35 (5.1, 11.41) & 14.47 (13.23, 15.7) & 7.47 (4.7, 10.2)\\
    \textbf{Gender} & & & \\
        \hspace{5mm}\textbf{Male} & 42024 (52.66\%) & 3997 (39.95\%) & 5509 (55.58\%)\\
        \hspace{5mm}\textbf{Female} & 37786 (47.34\%) & 6008 (60.05\%) & 4402 (44.42\%)\\
    \textbf{MCV} & 88.35 (75.28, 102.54) & 85.28 (68.1, 102.42) & 110.04 (105.0, 115.14)\\
    \textbf{Ferritin} & 263.51 (87.09, 423.78) & 297.77 (149.84, 442.46) & 301.19 (151.44, 451.82)\\
    \textbf{CRP} & 200.8 (101.85, 299.34) & 199.13 (99.52, 299.82) & 200.92 (103.09, 299.35)\\
    \textbf{TSAT} & 43.78 (19.29, 66.08) & 47.58 (26.44, 68.93) & 47.32 (26.08, 68.24)\\
    \textbf{Reticulocytes} & 290.86 (132.02, 442.36) & 305.54 (160.04, 454.44) & 306.87 (161.22, 453.58)\\
    \textbf{Haptoglobin} & 1.6 (0.31, 2.83) & 2.24 (1.11, 3.37) & 2.26 (1.11, 3.38)\\
    \textbf{Vitamin B12} & 778.98 (421.38, 1141.58) & 763.02 (385.25, 1135.72) & 750.62 (383.11, 1116.27)\\
    \textbf{Folate} & 13.39 (4.39, 21.37) & 15.58 (8.26, 22.75) & 3.49 (1.58, 1.91)\\
    \textbf{Sickle cells} & & & \\
        \hspace{5mm}\textbf{Positive} & 9917 (12.43\%) & 1 (0.01\%) & 0 (0\%)\\
        \hspace{5mm}\textbf{Negative} & 9901 (12.41\%) & 1 (0.01\%) & 0 (0\%)\\
  \hline
\end{tabular}

\bigskip

\begin{tabular}[]{@{}|l|l|l|l|@{}}
  \hline
    \bfseries Feature & \bfseries Unspecified anemia & \bfseries \makecell{Anemia of \\ chronic disease} & \bfseries Iron deficiency anemia \\\hline
         \textbf{Hemoglobin} & 7.45 (4.64, 10.23) & 7.5 (4.74, 10.22) &  7.46 (4.72, 10.26)\\
    \textbf{Gender} & & & \\
        \hspace{5mm}\textbf{Male} & 5443 (53.97\%) & 5358 (53.59\%) & 5504 (55.06\%) \\
        \hspace{5mm}\textbf{Female} & 4642 (46.03\%) & 4641 (46.41\%) &  4492 (44.94\%)\\
    \textbf{MCV} & 110.08 (105.08, 115.06) & 64.84 (57.36, 72.35) &  65.08 (57.65, 72.55)\\
    \textbf{Ferritin} & 300.31 (150.64, 452.08) & 315.07 (172.22, 457.8) &  50.3 (25.68, 74.87)\\
    \textbf{CRP} & 200.6 (102.33, 299.5) & 200.06 (100.79, 297.56) &  202.83 (104.47, 300.15)\\
    \textbf{TSAT} & 47.94 (26.95, 69.3) & 52.57 (34.63, 71.4) &  18.6 (9.42, 18.27)\\
    \textbf{Reticulocytes} & 305.35 (158.5, 455.09) & 305.82 (154.06, 456.43) &  303.39 (153.36, 449.26)\\
    \textbf{Haptoglobin} & 2.23 (1.08, 3.34) & 2.24 (1.11, 3.36) &  2.26 (1.16, 3.38)\\
    \textbf{Vitamin B12} & 853.83 (526.72, 1176.77) & 764.46 (397.62, 1141.15) &  765.02 (394.64, 1135.78)\\
    \textbf{Folate} & 15.83 (8.77, 22.86) & 15.67 (8.61, 22.77) &  15.7 (8.49, 22.91)\\
    \textbf{Sickle cells} &  &  &  \\
        \hspace{5mm}\textbf{Positive} & 0 (0\%) & 0 (0\%) & 0 (0\%) \\
        \hspace{5mm}\textbf{Negative} & 0 (0\%) & 0 (0\%) &  0 (0\%) \\
\hline
\end{tabular}

\bigskip
\begin{tabular}[]{@{}|l|l|l|l|@{}}
  \hline
    \bfseries Feature & \bfseries Other hemolytic anemia & \bfseries Aplastic anemia & \bfseries \makecell{Sickle cell \\ anemia }\\\hline
    \textbf{Hemoglobin}& 7.51 (4.8, 10.22) & 7.5 (4.73, 10.27) &  7.45 (4.65, 10.23)\\
    \textbf{Gender} & & & \\
        \hspace{5mm}\textbf{Male} & 5311 (53.65\%) & 5454 (54.55\%) &  5448 (54.94\%)\\
        \hspace{5mm}\textbf{Female} & 4589 (46.35\%) & 4544 (45.45\%) &  4468 (45.06\%)\\
    \textbf{MCV} & 89.96 (85.0, 95.0) & 89.96 (84.97, 95.0) &  90.04 (85.01, 95.09)\\
    \textbf{Ferritin} & 303.36 (151.75, 452.54) & 305.11 (155.45, 457.96) &  300.11 (153.26, 446.19)\\
    \textbf{CRP} & 201.48 (100.56, 300.1) & 200.22 (99.42, 298.09) &  201.26 (104.72, 299.5)\\
    \textbf{TSAT} & 46.65 (25.26, 67.93) & 47.86 (27.54, 69.02) &  47.55 (26.4, 68.74)\\
    \textbf{Reticulocytes} & 374.0 (262.54, 485.16) & 79.84 (45.72, 114.36) &  376.98 (267.09, 487.13)\\
    \textbf{Haptoglobin} & 0.25 (0.13, 0.38) & 2.27 (1.15, 3.39) &  0.25 (0.12, 0.37)\\
    \textbf{Vitamin B12} & 764.31 (389.02, 1138.31) & 755.54 (385.21, 1123.64) &  757.31 (380.99, 1131.96)\\
    \textbf{Folate} & 15.48 (8.44, 22.51) & 15.56 (8.46, 22.6) &  15.7 (8.77, 22.67)\\
    \textbf{Sickle cells} &  &  &  \\
        \hspace{5mm}\textbf{Positive} & 0 (0\%) & 0 (0\%) & 9916 (100\%) \\
        \hspace{5mm}\textbf{Negative} & 9900 (100\%) & 0 (0\%) & 0 (0\%) \\
    \hline
\end{tabular}

\caption{Summary descriptive statistics of the synthetic dataset showing the mean and interquartile range (in parentheses) of the features. Gender and sickle cells, which are binary variables, are described using the sample number and the percentage.}
\label{tab:syn_summary_stats}

\end{table}

\subsection{Real-World Dataset}\label{apd:rwd_dataset}
\begin{figure}[htbp]
    \centering
    \includegraphics[width=\linewidth]{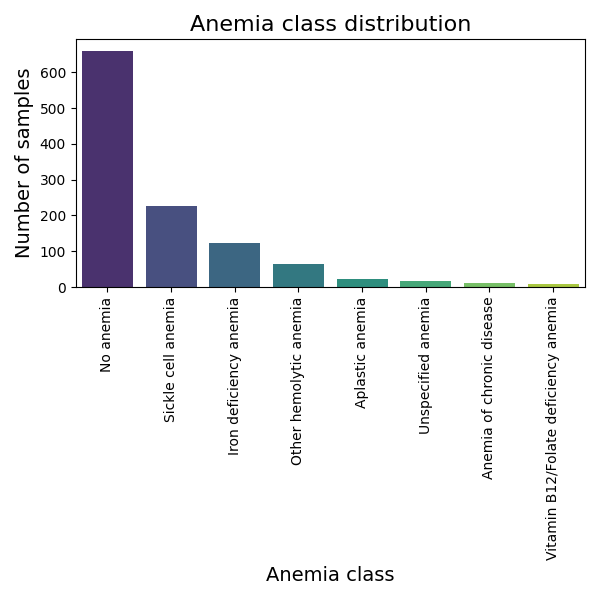}
    \caption{Class distribution in the real-world dataset.}
    \label{fig:rwd_class_distribution}
\end{figure}

\begin{figure}[htbp]
    \centering
    \includegraphics[width=\linewidth]{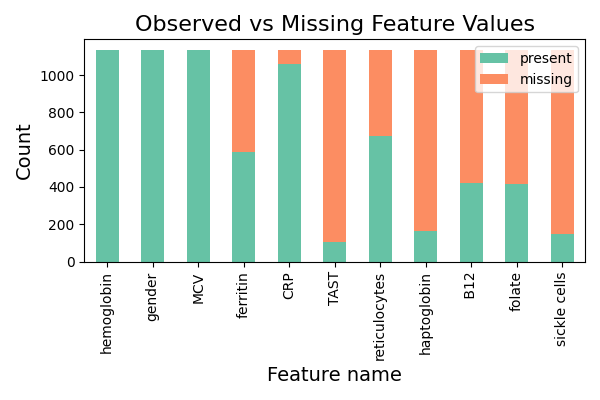}
    \caption{Observed \textit{vs.} missing values for each feature in the real-world dataset.}
    \label{fig:rwd_missing}
\end{figure}

Figure~\ref{fig:rwd_class_distribution} shows the class distribution of the real-world dataset, Figure~\ref{fig:rwd_missing} depicts the ratio of observed versus missing values for each feature, Table \ref{tab:rwd_instance} illustrates an example instance in the dataset, and Table \ref{tab:rwd_summary_stats} shows the summary statistics of thes dataset.

\begin{table}[htbp]
\centering  
    \begin{tabular}[]{@{}|l|l|@{}}
    \hline
         \bfseries Feature & \bfseries Value \\\hline
         \textbf{Hemoglobin} & 8.2 g/100mL \\
         \textbf{Gender} & Female \\
         \textbf{MCV} & 80.0 fL \\
         \textbf{Ferritin} &  16.0 $\mu$g/L\\
         \textbf{CRP} & 7.2 mg/L\\
         \textbf{TSAT} & - \\
         \textbf{Reticulocytes} & 66.00 G/L \\
         \textbf{Haptoglobin} & 2.43 g/L \\
         \textbf{Vitamin B12} & 214.00 ng/L\\
         \textbf{Folate} & 4.6 $\mu$g/L\\
         \textbf{Sickle cells} & - \\         
         \textbf{Label} & Iron deficiency anemia\\\hline
    \end{tabular}
    \caption{An instance in the real-world dataset.}
    \label{tab:rwd_instance}
\end{table}

\begin{table}[htbp]
    \fontsize{7pt}{7pt}\selectfont
    \centering
    
    \begin{tabular}[]{@{}|l|l|l|l|@{}}

  \hline
    \bfseries Feature & \bfseries All Classes & \bfseries No anemia & \bfseries \makecell{Vitamin B12/Folate \\ deficiency anemia} \\ \hline
    \textbf{Hemoglobin} & 11.42 (9.20, 13.7) & 13.33 (12.4, 14.4) & 6.37 (4.7, 6.68)\\
    \textbf{Gender} & & & \\
        \hspace{5mm}\textbf{Male} & 517 (45.55\%) & 321 (48.71\%) & 7 (70.0\%) \\
        \hspace{5mm}\textbf{Female} & 618 (54.45\%) & 338 (51.29\%) &  3 (30.0\%)\\
    \textbf{MCV} & 84.96 (81.0, 91.0) & 87.01 (84.0, 91.0) &  111.5 (99.75, 125.25)\\
    \textbf{Ferritin} & 410.21 (28.0, 391.0) & 410.51 (71.5, 410.0) &  364.44 (319.0, 476.0)\\
    \textbf{CRP} & 47.02 (2.6, 57.95) & 55.07 (2.7, 75.7) &  28.45 (0.9, 22.02)\\
    \textbf{TSAT} & 28.57 (13.0, 42.25) & 30.81 (14.0, 43.25)  & 44.0 (4.0, 83.0) \\
    \textbf{Reticulocytes} & 117.04 (44.0, 156.25) & 58.62 (36.0, 73.0) & 28.8 (12.0, 44.25)\\
    \textbf{Haptoglobin} & 1.22 (0.09, 1.98) & 1.69 (0.82, 2.33) &  0.23 (0.01, 0.05)\\
    \textbf{Vitamin B12} & 375.1 (203.0, 452.0) & 399.76 (213.0, 486.25) &  78.7 (25.0, 61.25)\\
    \textbf{Folate} & 9.59 (5.1, 11.02) & 8.33 (5.1, 9.9) &  7.74 (2.6, 9.57)\\
    \textbf{Sickle cells} &  &  &  \\
        \hspace{5mm}\textbf{Positive} & 147 (12.95\%) & 0 (0\%)& 0 (0\%) \\
        \hspace{5mm}\textbf{Negative}  & 0 (0\%) & 0 (0\%) & 0 (0\%) \\
  \hline
\end{tabular}

\bigskip

\begin{tabular}[]{@{}|l|l|l|l|@{}}
  \hline
    \bfseries Feature & \bfseries Unspecified anemia & \bfseries \makecell{Anemia of \\ chronic disease} & \bfseries Iron deficiency anemia \\\hline
         \textbf{Hemoglobin} & 6.76 (5.45, 8.0) & 8.07 (6.70, 9.80) & 7.95 (6.0, 10.22)\\
    \textbf{Gender} & & & \\
        \hspace{5mm}\textbf{Male} & 6 (37.50\%) & 7 (63.64\%) & 30 (24.19\%)\\
        \hspace{5mm}\textbf{Female} & 10 (62.50\%) & 4 (36.36\%) & 94 (75.81\%)\\
    \textbf{MCV} & 89.25 (77.25, 100.0) & 82.09 (71.5, 92.5) & 73.5 (64.75, 81.25)\\
    \textbf{Ferritin} & 408.19 (35.25, 403.25) & 194.88 (22.75, 307.5) & 42.97 (4.0, 19.0)\\
    \textbf{CRP} & 30.69 (2.7, 28.62) & 81.87 (17.4, 111.4) & 13.76 (1.18, 13.68)\\
    \textbf{TSAT} & 39.67 (16.75, 52.25) & - & 10.8 (3.0, 14.75)\\
    \textbf{Reticulocytes} & 68.44 (35.0, 93.5) & 74.36 (35.5, 102.0) & 59.09 (39.75, 73.25)\\
    \textbf{Haptoglobin} & 1.69 (0.83, 3.12) & 2.22 (2.04, 2.33) & 1.6 (1.01, 2.04)\\
    \textbf{Vitamin B12} & 315.4 (208.5, 411.0) & (220.5, 387.5) & (187.75, 435.0)\\
    \textbf{Folate} & 8.18 (4.8, 9.95) & 8.81 (5.9, 11.85) & 9.21 (5.1, 10.4)\\
    \textbf{Sickle cells} &  &  &  \\
        \hspace{5mm}\textbf{Positive}  & 0 (0\%) & 0 (0\%) & 0 (0\%) \\
        \hspace{5mm}\textbf{Negative}  & 0 (0\%) & 0 (0\%) & 0 (0\%) \\
\hline
\end{tabular}

\bigskip
\begin{tabular}[]{@{}|l|l|l|l|@{}}
  \hline
    \bfseries Feature & \bfseries Other hemolytic anemia & \bfseries Aplastic anemia & \bfseries \makecell{Sickle cell \\ anemia }\\\hline
    \textbf{Hemoglobin} & 9.85 (7.2, 11.7) & 8.92 (8.15, 10.32) & 9.22 (7.9, 10.2)\\
    \textbf{Gender} & & & \\
        \hspace{5mm}\textbf{Male} & 20 (30.77\%)  & 14 (58.33\%) & 112 (49.56\%)\\
        \hspace{5mm}\textbf{Female} & 45 (69.23\%) & 10 (41.67\%) & 114 (50.44\%)\\
    \textbf{MCV} & 88.1 (77.0, 101.5) & 89.92 (87.0, 95.0) & 82.51 (73.0, 90.0)\\
    \textbf{Ferritin} & 758.22 (102.0, 460.0) & 1104.27 (543.0, 1622.0) & 588.56 (60.5, 446.75)\\
    \textbf{CRP} & 24.06 (2.7, 25.7) & 109.77 (26.85, 167.52) & 40.51 (3.8, 35.7)\\
    \textbf{TSAT} & 27.0 (19.0, 33.0) & 51.5 (34.25, 68.75) & 31.96 (17.5, 38.0)\\
    \textbf{Reticulocytes} & 149.21 (62.75, 233.0) & 32.61 (12.0, 45.5) & 213.19 (122.25, 278.5)\\
    \textbf{Haptoglobin} & 0.43 (0.01, 0.48) & 1.58 (0.88, 1.78) & 0.36 (0.04, 0.29)\\
    \textbf{Vitamin B12} & 364.24 (214.0, 452.0) & 453.8 (240.75, 580.5) & 351.21 (180.0, 393.5) \\
    \textbf{Folate} & 10.76 (5.7, 12.4) & 9.03 (5.12, 9.6) & 16.91 (8.85, 25.2)\\
    \textbf{Sickle cells} &  &  &  \\
        \hspace{5mm}\textbf{Positive} & 0 (0\%) & 0 (0\%) & 147 (65.04\%)  \\
        \hspace{5mm}\textbf{Negative} & 0 (0\%) & 0 (0\%) & 0 (0\%) \\
     \hline
\end{tabular}

\caption{Summary descriptive statistics of the real-world dataset showing the mean and interquartile range (in parentheses) of the features. Gender and sickle cells, which are binary variables, are described using the sample number and the percentage.}
\label{tab:rwd_summary_stats}
\end{table}

\section{More Results}\label{apd:more_results}
Table \ref{tab:from_scratch} presents the results of the RWD-trained models,
Table \ref{tab:as_is} shows the results of the synthetic dataset-trained models without modification, and Table \ref{tab:finetuned} shows the results of the synthetic-trained models fine-tuned using the real-world dataset.

\begin{table}[!htbp]
\centering
\resizebox{\textwidth}{!}{ 
\begin{tabular}{|l|l|l|l|l|l|}
\hline
\textbf{Model} & \textbf{Goal metric} & \textbf{Accuracy} & \textbf{F1} & \textbf{ROC-AUC} & \textbf{MEL}\\ 
\hline
\multirow{3}{*}{\makecell{Dueling\\ DQN-PER}} & Accuracy & 77.02 ± 3.03 & 45.45 ± 6.37 & 73.34 ± 5.21 & 3.40 ± 0.23\\
\cline{2-6}
& F1 & \textbf{78.86 ± 1.73} & \textbf{48.22 ± 5.79} & \textbf{74.95 ± 3.59} & 3.29 ± 0.31 \\
\cline{2-6}
& ROC-AUC & 77.98 ± 3.05 & 45.08 ± 6.06 & 72.87 ± 3.90 & 3.27 ± 0.23\\
\hline
\multirow{3}{*}{\makecell{Dueling\\ DDQN-PER}} & Accuracy & 75.79 ± 2.69 & 40.15 ± 6.69 & 67.88 ± 4.17 & 3.34 ± 0.33\\
\cline{2-6}
& F1 & 77.37 ± 2.33 & 42.59 ± 5.42 & 70.26 ± 2.23 & \textbf{3.09 ± 0.29}\\
\cline{2-6}
& ROC-AUC & 77.28 ± 2.13 & 44.05 ± 3.64 & 70.47 ± 2.73 & 3.17 ± 0.30\\
\hline
\end{tabular}
}
\caption{The mean performance of the RWD-trained models, based on the selected goal metric (Accuracy or F1 or ROC-AUC).}
\label{tab:from_scratch}%
  
\end{table} 

\begin{table}[!htbp]
\centering
\resizebox{\textwidth}{!}{ 
  
    \begin{tabular}{|l|l|l|l|l|}
    \hline
    \bfseries Model & \bfseries Accuracy & \bfseries F1 & \bfseries ROC-AUC & \bfseries MEL\\\hline
        Learned decision tree & \textbf{66.67 ± 0.00} & 34.31 ± 1.04 & 68.56 ± 0.04 & 7.80 ± 0.00 \\ \hline
        Dueling DQN-PER & 57.02 ± 0.00 & 33.31 ± 0.00  &  69.91 ± 0.00  & \textbf{3.16 ± 0.00}  \\ \hline
        Dueling DDQN-PER & 57.89 ± 0.00  & \textbf{35.58 ± 0.00}  &  \textbf{73.22 ± 0.00} &  3.26 ± 0.00 \\ \hline
    \end{tabular}
    }
\caption{Performance of synthetic dataset-trained models on the RWD test set with metrics presented as the average and standard deviation over 10 runs.}
\label{tab:as_is}%
\end{table}

\begin{table}[!htbp]
\centering
\resizebox{\textwidth}{!}{ 
\begin{tabular}{|l|l|l|l|l|l|}
\hline
\textbf{Model} & \textbf{Goal metric} & \textbf{Accuracy} & \textbf{F1} & \textbf{ROC-AUC} & \textbf{MEL}\\ 
\hline
\multirow{3}{*}{\makecell{Dueling\\ DQN-PER}} & Accuracy & \textbf{81.58 ± 0.00} &  51.75 ± 0.00  & 74.34 ± 0.00  & 3.02 ± 0.00 \\
\cline{2-6}
& F1 & 78.07 ± 0.00 & 53.03 ± 0.00 & \textbf{77.88 ± 0.00}& 3.13 ± 0.00\\
\cline{2-6}
& ROC-AUC & 77.19 ± 0.00 & \textbf{63.48} ± 0.00  & 77.67 ± 0.00  & 3.19 ± 0.00 \\
\hline
\multirow{3}{*}{\makecell{Dueling\\ DDQN-PER}} & Accuracy & 73.68 ± 0.00 & 48.93 ± 0.00  & 69.15 ± 0.00  & 3.85 ± 0.00 \\
\cline{2-6}
& F1 & 72.81 ± 0.00 & 33.19 ± 0.00  & 64.78 ± 0.00  & 3.76 ± 0.00 \\
\cline{2-6}
& ROC-AUC & 79.82 ± 0.00 & 44.30 ± 0.00  & 72.09 ± 0.00  & \textbf{2.90 ± 0.00} \\
\hline
\end{tabular}
}
\caption{The mean performance of the fine-tuned synthetic dataset-trained models, based on the selected goal metric (Accuracy or F1 or ROC-AUC).}
\label{tab:finetuned}%

\end{table}

\section{The Pathways}\label{apd:pathways}
\begin{figure}[htbp]
    \centering
    \includegraphics[width=\textwidth]{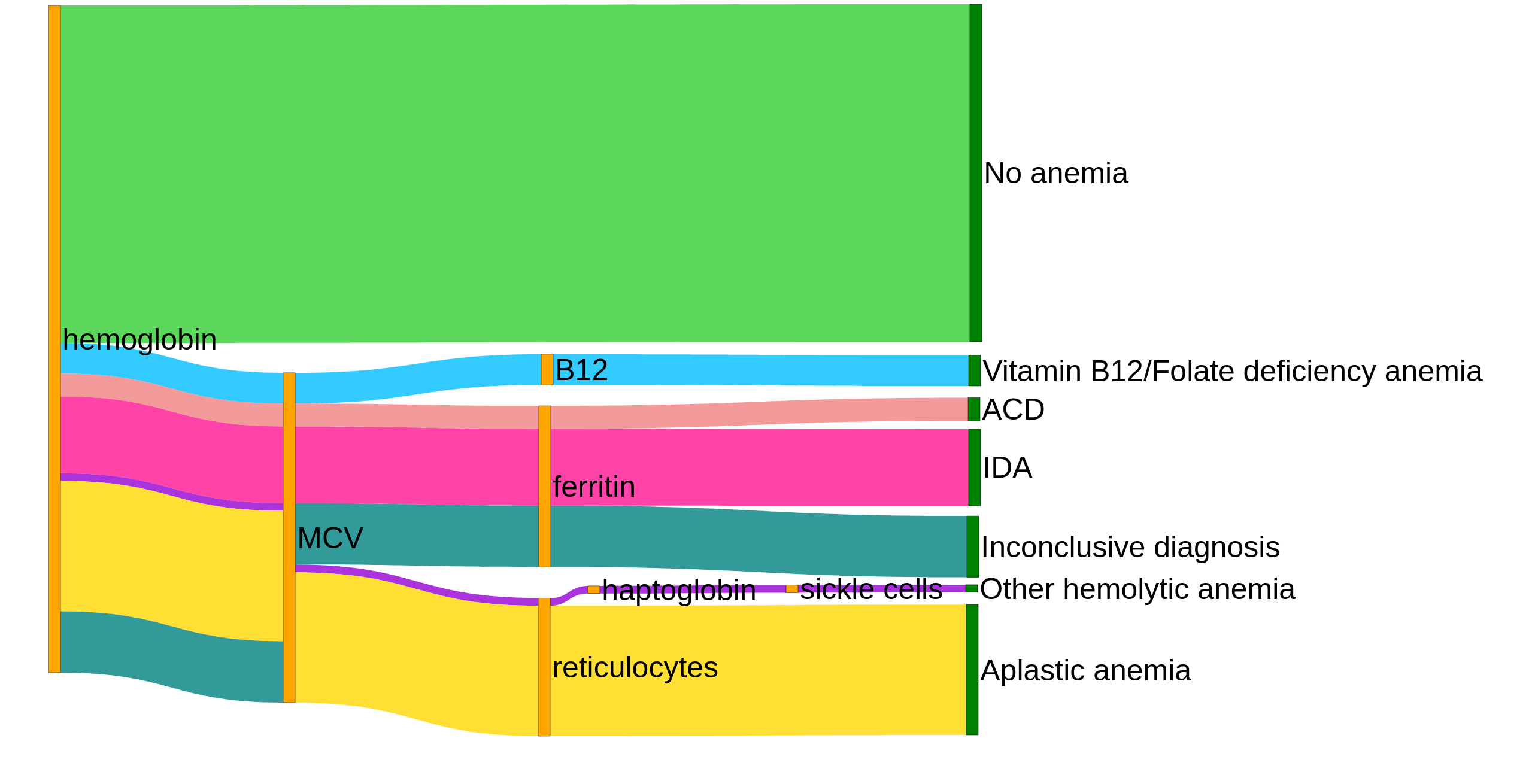}
    \caption{Most common diagnostic pathways from the expert-defined DT.}
    \label{fig:modified_dt_pathway}
\end{figure}

\begin{figure}[htbp]
    \centering
    \includegraphics[width=\textwidth]{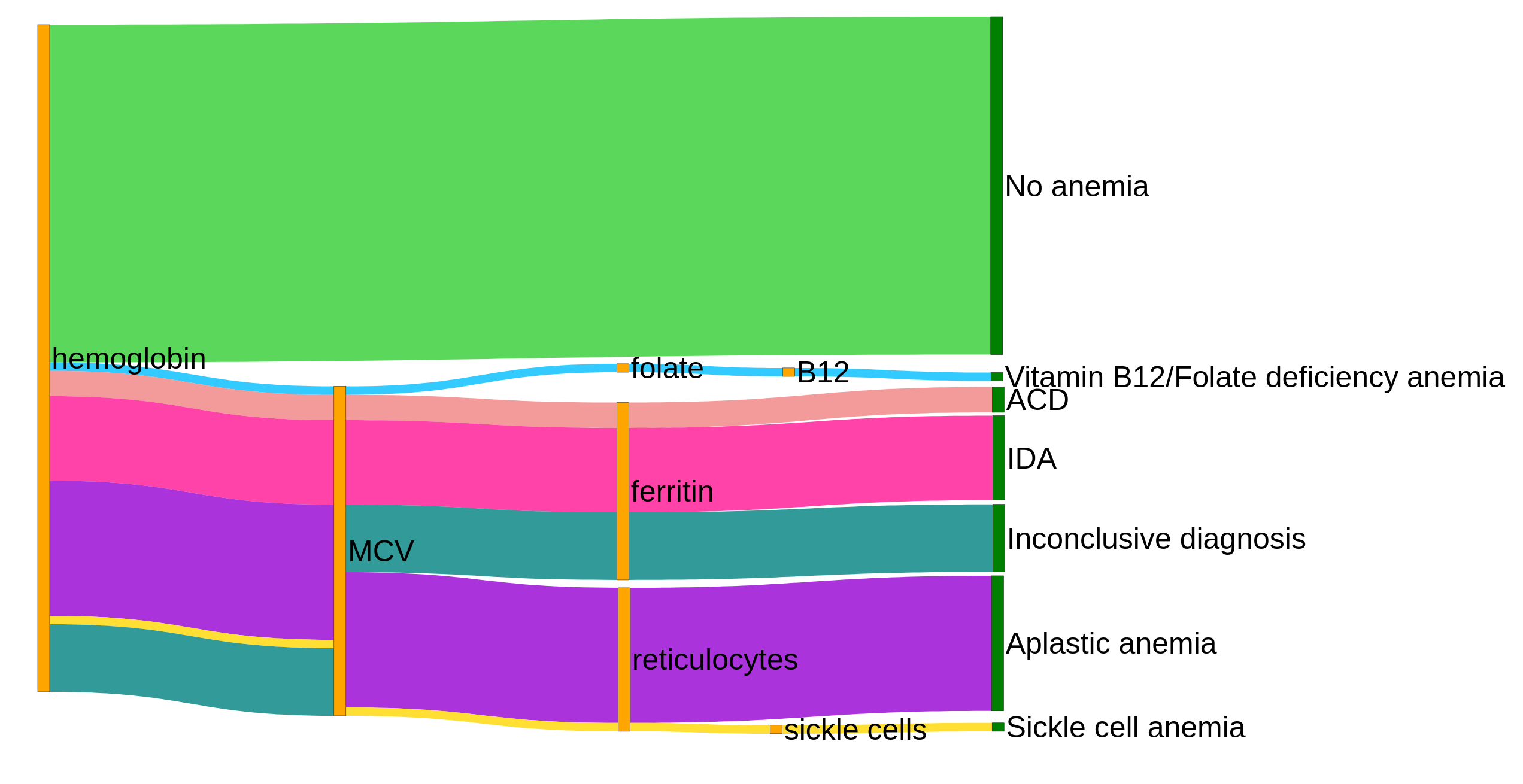}
    \caption{Most common diagnostic pathways from the best-performing synthetic dataset-trained DQN model.}
    \label{fig:pretrained_pathway}
\end{figure}

\begin{figure}[htbp]
    \centering
    \includegraphics[width=\textwidth]{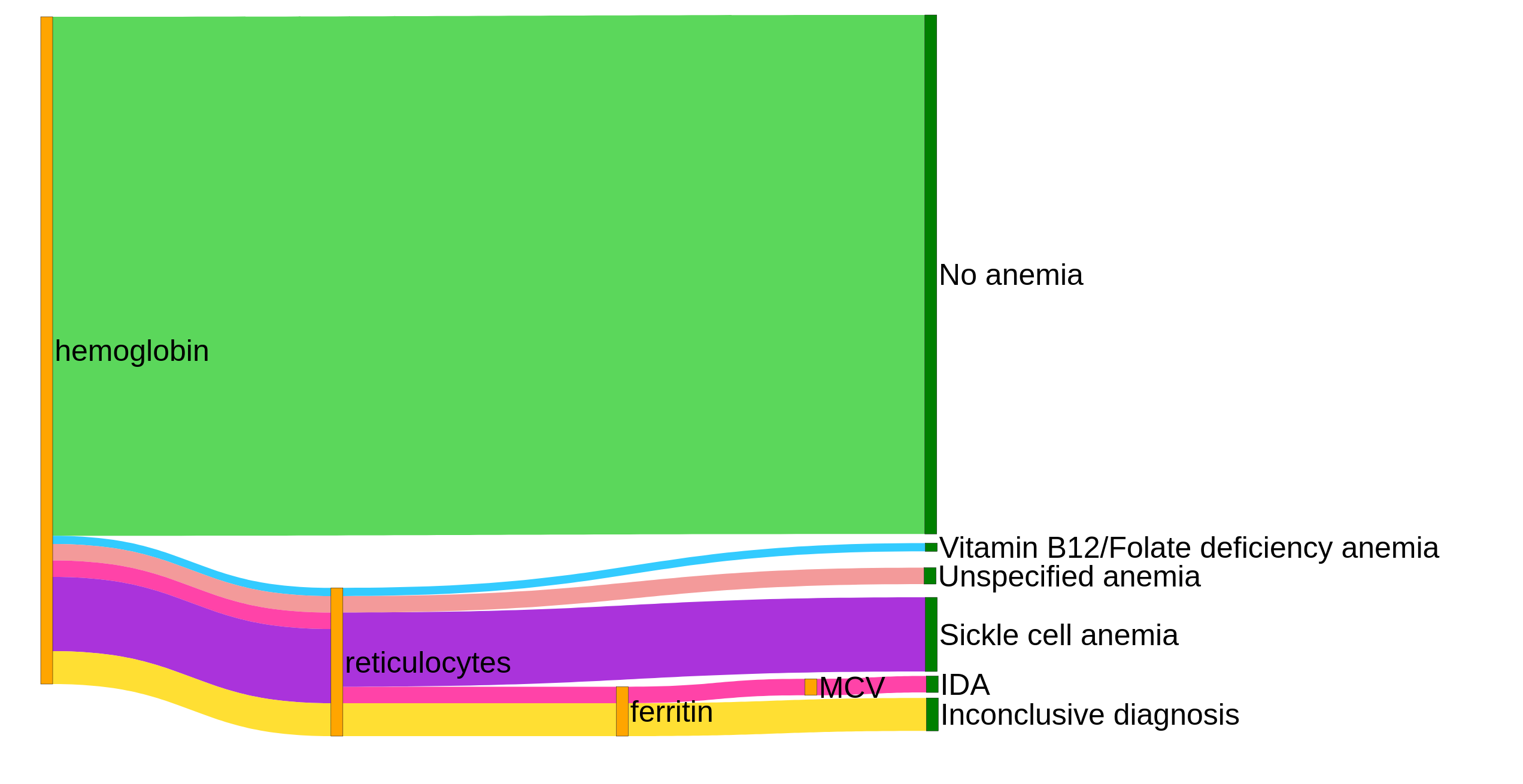}
    \caption{Most common diagnostic pathways from the best-performing RWD-trained DQN model.}
    \label{fig:from_scratch_pathway}
\end{figure}

\begin{figure}[htbp]
    \centering
    \includegraphics[width=\textwidth]{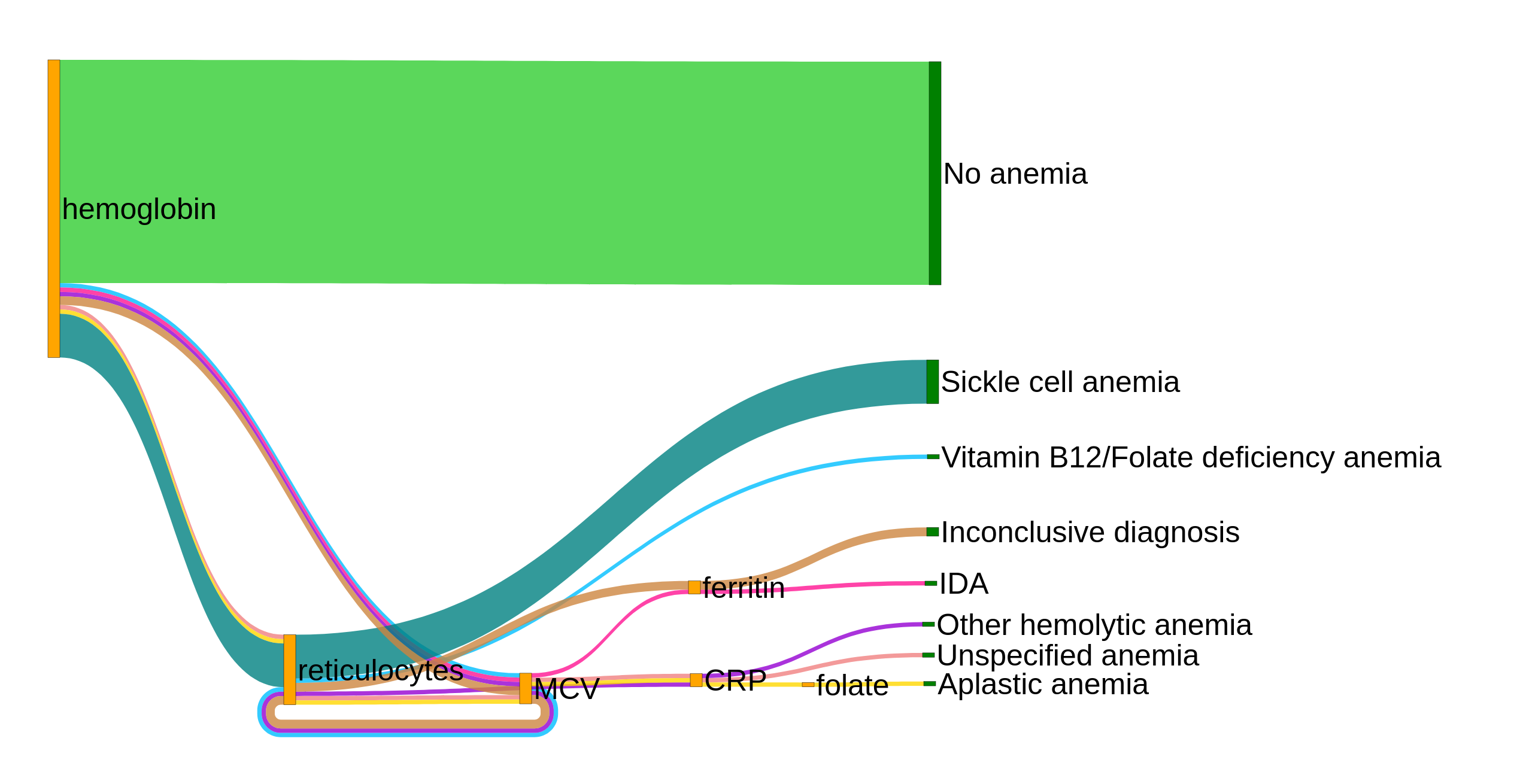}
    \caption{Most common diagnostic pathways from the best-performing fine-tuned DQN model.}
    \label{fig:finetuned_pathway}
\end{figure}

Figures \ref{fig:modified_dt_pathway} to \ref{fig:finetuned_pathway} 
 illustrate the most common pathway for each diagnosed anemia class as generated by the expert-defined DT, the best-performing synthetic dataset-trained DQN model, the best-performing RWD-trained DQN model, and the the best-performing fine-tuned DQN model in terms of the F1 score, respectively. In these figures, orange nodes represent the patient features requested by the model, ordered from left to right, while dark green nodes indicate the final diagnoses. Each anemia class is represented by a different color, and the thickness of each link is proportional to its support, \textit{i.e.} the number of patients who have it in their pathway.

\section{Computing Time}
In Table \ref{tab:computing_time}, the time taken to train each model (training time), as well as, the time taken to generate a diagnosis pathway and a diagnosis (testing time) for the DQN models and the SOTA models respectively for the RWD test set are shown.

\begin{table}[!ht]
\centering

    \begin{tabular}{|p{4.2cm}|p{4cm}|p{4cm}|}
    \hline
    \bfseries Model & \bfseries Training time & \bfseries Testing time\\
    \hline
         Random Agent & N/A & 96.7 ms ± 545 µs\\
         Tree-based agent & N/A & 98.7 ms ± 497 µs\\
         RWD-trained decision tree &  2.35 ms ± 43.2 µs &  394 µs ± 1.07 µs\\ 
         Random Forest & 81.3 ms ± 163 µs  &  4.28 ms ± 24 µs\\
         XGBoost & 535 ms ± 100 ms &  6.23 ms ± 3.16 ms\\ 
         
         FFNN & 21.8 s ± 532 ms & 1.12 ms ± 17.3 µs\\ 
         Fine-tuned Dueling DQN-PER & 6h 13min 25s ± 24.7 s &  158 ms ± 3.51 ms\\ 
    \hline
    \end{tabular}
\caption{The computing time to train the models and to generate pathways and/or diagnoses for the RWD test set. Results are mean and standard deviation over 5 runs.}

\label{tab:computing_time}
  
\end{table}

\end{document}